\RequirePackage{fix-cm}
\documentclass[twocolumn]{svjour3}          
\smartqed  
\usepackage{graphicx}
\usepackage{mathptmx}
\usepackage{subfigure,multirow,array,amssymb,amsmath,natbib}
\usepackage[misc,geometry]{ifsym} 
\usepackage[colorlinks,citecolor=blue]{hyperref}

\newcommand{\tabincell}[2]{\begin{tabular}{@{}#1@{}}#2\end{tabular}}

\newcommand{\highlight}[1]{\textcolor{black}{#1}}

\journalname{International Journal of Computer Vision}
\begin{document}

\title{J\texorpdfstring{$\hat{\text{A}}$}{Lg}A-Net: Joint Facial Action Unit Detection and Face Alignment via Adaptive Attention\thanks{Zhiwen~Shao and Zhilei~Liu contributed equally to this work.}
}


\author{Zhiwen~Shao$^{1,2,3}$ \and
        Zhilei~Liu$^4$ \and
        Jianfei~Cai$^5$ \and
        Lizhuang~Ma$^{3,6}$
}


\institute{Zhiwen~Shao (\Letter) \at
          \email{zhiwen\_shao@cumt.edu.cn}
          \and
          $^1$School of Computer Science and Technology, China University of Mining and Technology, Xuzhou 221116, China\\
          $^2$Engineering Research Center of Mine Digitization, Ministry of Education of the People’s Republic of China, Xuzhou 221116, China\\
          $^3$Department of Computer Science and Engineering, Shanghai Jiao Tong University, Shanghai 200240, China\\
          $^4$College of Intelligence and Computing, Tianjin University, Tianjin 300072, China\\
          $^5$Faculty of Information Technology, Monash University, Clayton 3800, VIC, Australia\\
          $^6$School of Computer Science and Technology, East China Normal University, Shanghai 200062, China
}

\date{Received: date / Accepted: date}

\maketitle

\begin{abstract}
Facial action unit (AU) detection and face alignment are two highly correlated tasks, since facial landmarks can provide precise AU locations to facilitate the extraction of meaningful local features for AU detection. However, most existing AU detection works handle the two tasks independently by treating face alignment as a preprocessing, and often use landmarks to predefine a fixed region or attention for each AU. In this paper, we propose a novel end-to-end deep learning framework for joint AU detection and face alignment, which has not been explored before. In particular, multi-scale shared feature is learned firstly, and high-level feature of face alignment is fed into AU detection. Moreover, to extract precise local features, we propose an adaptive attention learning module to refine the attention map of each AU adaptively. Finally, the assembled local features are integrated with face alignment feature and global feature for AU detection. Extensive experiments demonstrate that our framework (i) significantly outperforms the state-of-the-art AU detection methods on the challenging BP4D, DISFA, GFT and BP4D+ benchmarks, (ii) can adaptively capture the irregular region of each AU, (iii) achieves competitive performance for face alignment, and (iv) also works well under partial occlusions and non-frontal poses. The code for our method is available at \textit{https://github.com/ZhiwenShao/ PyTorch-JAANet}.
\keywords{Joint learning \and Facial AU detection \and Face alignment \and Adaptive attention learning}
\end{abstract}

\section{Introduction}

Facial action unit (AU) detection and face alignment are two important face analysis tasks in the fields of computer vision and affective computing~\citep{corneanu2016survey,TAC2017Pantic}. In most of face related tasks, face alignment~\citep{kazemi2014one,zhang2016learning,shao2019deep} is usually employed to localize certain distinctive facial locations, namely landmarks, to define the facial shape or expression appearance. Facial AUs refer to a unique set of basic facial actions at certain facial locations defined by Facial Action Coding System (FACS)~\citep{ekman1978facial,ekman2002facial}, which is one of the most comprehensive and objective systems for describing facial expressions. Considering AU detection and face alignment are coherently related to each other, they should be beneficial for each other if putting them in a joint framework. However, in literature it is rare to see such joint study of the two tasks.

In some previous AU detection studies~\citep{gudi2015deep,zhao2016deep,chu2017learning}, facial landmarks are only used to align faces into a common reference face, so that the extracted features from each face correspond to the same semantic locations. Since landmarks can also provide precise AU locations, recent works pay more attention to extracting AU-related features from regions of interest (ROIs) centered around the associated landmarks. For example, \cite{li2018eac,li2017action} proposed a deep learning based approach named EAC-Net for AU detection by enhancing and cropping the ROIs with landmark information. However, they just treated face alignment as a preprocessing. 
\cite{wu2016constrained} tried to exploit face alignment and AU detection simultaneously with a cascade regression framework, which is a pioneering work for the joint study of the two tasks. However, this cascade regression method only uses handcrafted features and is not based on the prevailing deep learning technology, which limits its performance.

In addition to EAC-Net~\citep{li2018eac} which predefines the ROI of each AU with a fixed size and a fixed attention distribution, a few works also adopt the attention mechanism. \cite{sanchez2018joint} used a predefined Gaussian distribution to generate an attention map for each AU, in which the amplitude and size of the Gaussian distribution are determined by the AU intensity. However, these methods cannot adapt to various AUs with irregular shapes and transformations. Recently, \cite{shao2019facial} directly learned spatial attentions of AUs without the prior landmark knowledge. Although this work can find the irregular AU regions, some irrelevant regions are also captured.

To tackle the above limitations, we propose a novel deep learning based joint AU detection and face alignment framework to exploit the strong correlation of the two tasks. In particular, multi-scale feature shared by the two tasks is lear-ned firstly, and high-level feature of face alignment is extracted and fed into AU detection. Moreover, to extract precise local features, we propose an adaptive attention learning module to refine the attention map of each AU adaptively, which is initially specified by the predicted facial landmarks. Finally, the assembled local features are integrated with face alignment feature and global feature for AU detection. In the adaptive attention learning module, each AU has an independent branch to refine its attention map under the supervision of its local AU detection loss. Besides, the face alignment feature and the global feature supplement other useful information on top of the assembled local features. The entire framework is end-to-end without any post-processing operation, and all the modules are optimized jointly.

The contributions of this paper are threefold:
\begin{itemize}
    \item We propose an end-to-end multi-task deep learning framework for joint facial AU detection and face alignment. To the best of our knowledge, jointly modeling these two tasks with deep neural networks has not been done before.
    \item With the aid of face alignment results, an adaptive attention network is learned to determine the attention distribution of the ROI of each AU.
    \item We conduct extensive experiments on benchmarks, where our proposed joint framework significantly outperforms the state-of-the-art AU detection methods, can adaptively capture the irregular region of each AU, achieves competitive performance for face alignment, and also works well under partial occlusions and non-frontal poses.
\end{itemize}

In comparison to the earlier conference version~\citep{shao2018deep} of this work, we introduce the new local AU detection loss in Sec.~\ref{ssec:au_det} to generalize the original idea of the back-propagation enhancement. Specifically, we show that the local AU detection loss is a more effective way to supervise the refinement of attention maps so as to extract more precise local features. We also remove the constraint on the differences of the attention maps before and after the refinement, which reduces the restrictions from predefined attention maps and thus facilitates the adaptive learning of attentions. 
With these improvements, our framework becomes more general and achieves better AU detection performance. Aside from the changes in methodology, this extension also supplements the comparisons on challenging GFT~\citep{girard2017sayette} and BP4D+~\citep{zhang2016multimodal} benchmarks, as well as the results under partial occlusions and non-frontal poses. \highlight{We name our framework as \textbf{J$\hat{\text{A}}$A-Net} because of \textit{Joint} learning and \textit{Adaptive Attention}, in which ``$\hat{\text{A}}$'' corresponds to ``adaptive'', considering we improve the adaptive attention learning of JAA-Net in our earlier conference version~\citep{shao2018deep}.}


\section{Related Work}

Our proposed framework is closely related to existing landmark aided facial AU detection methods as well as joint facial AU detection and face alignment methods, since we combine both AU detection models and face alignment models.

\subsection{Landmark Aided Facial AU Detection}

The preprocessing step in most of the previous facial AU recognition works is to detect and align faces with the help of face detection and face alignment methods~\citep{TAC2017Pantic}. Considering it is robust to measure the landmark-based geometry changes, \cite{benitez2016emotionet} proposed an approach to fuse the geometry and local texture information for AU detection, in which the geometry information is obtained by measuring the normalized facial landmark distances and the angles of Delaunay mask formed by the landmarks. \cite{valstar2006fully} analyzed Gabor wavelet features near $20$ facial landmarks, and these features were then selected and classified by Adaboost and SVM classifiers for AU detection. \cite{zhao2015joint,zhao2016joint} proposed a Joint Patch and Multi-Label Learning (JPML) method for facial AU detection by taking into account both patch learning and multi-label learning, in which the local regions of AUs are defined as patches centered around the facial landmarks obtained using IntraFace~\citep{de2015intraface}. Recently, \cite{li2018eac,li2017action} proposed the EAC-Net for facial AU detection by enhancing and cropping the predefined ROI of each AU. All the ROIs with central locations specified by landmarks have a fixed size and a fixed attention distribution.

All these researches demonstrate the effectiveness of utilizing facial landmarks on feature extraction for AU detection task. However, they all treat face alignment as a single and independent task and make use of the existing well-designed facial landmark detectors.

\subsection{Joint Facial AU Detection and Face Alignment}

\highlight{As a task belonging to facial expression recognition, facial AU detection has a strong correlation with face alignment. The correlation between the two tasks can be exploited to help each other.} 

\highlight{On one hand, the correlation has been leveraged in several face alignment works.} For example, \cite{wu2017simultaneous} combined the tasks of face alignment, head pose estimation, and expression related facial deformation analysis using a cascade regression framework. \cite{zhang2014facial,zhang2016learning} proposed a Tasks-Constrained Deep Convolutional Network (TCDCN) to optimize the feature map shared between face alignment and other heterogeneous but subtly correlated tasks, e.g. head pose estimation and the inference of facial attributes including expression. \cite{ranjan2017hyperface} proposed a deep multi-task learning framework named HyperFace for simultaneous face detection, face alignment, pose estimation, and gender recognition. All these works demonstrate that related tasks such as facial expression recognition are beneficial for face alignment. However, in TCDCN and HyperFace, face alignment and other tasks are just simply integrated with the first several layers shared. In contrast, besides sharing feature layers, our proposed J$\hat{\text{A}}$A-Net also feeds high-level representations of face alignment into AU detection, and utilizes the estimated landmarks for the initialization of the adaptive attention learning.

\highlight{On the other hand, the correlation can also contribute to facial AU detection.} However, the interaction of the two tasks is usually one way in the aforementioned methods, i.e. facial landmarks are used to extract features for AU detection. Instead of treating face alignment independently, \cite{li2013simultaneous} proposed a hierarchical framework with Dynamic Bayesian Network to capture the joint local relationship between facial landmark tracking and facial AU recognition. However, this framework requires an offline facial activity model construction and an online facial motion measurement and inference, and only local dependencies between facial landmarks and AUs are considered. Inspired by~\citep{li2013simultaneous}, \cite{wu2016constrained} tried to exploit global AU relationship, global facial shape patterns, and global dependencies between AUs and landmarks with a cascade regression framework, which is a pioneering work for the joint process of the two tasks. In contrast with these conventional methods using handcrafted local appearance features, we employ an end-to-end deep framework for joint learning of facial AU detection and face alignment. Moreover, we develop a deep adaptive attention learning method to explore the feature distributions of different AUs in different ROIs specified by the predicted facial landmarks.

\begin{figure*}
\centering\includegraphics[width=\linewidth]{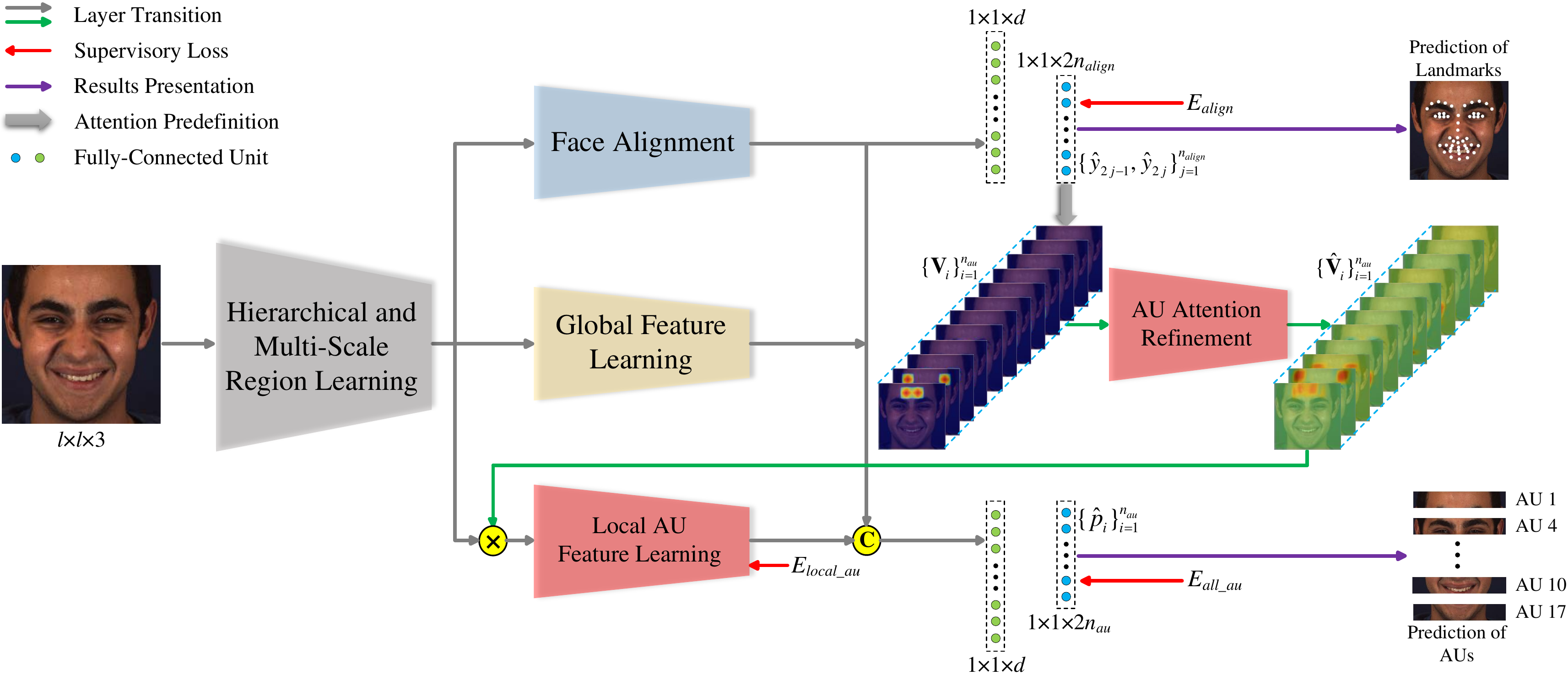}
\caption{The architecture of our J$\hat{\text{A}}$A-Net framework, where the submodules, AU attention refinement and local AU feature learning (in red), compose the adaptive attention learning module. The predefined and refined attention maps are overlaid on the input image for a better view. ``C'' denotes concatenation of feature map channels, and ``$\times$'' denotes element-wise multiplication of each feature map channel and the attention map.}
\label{fig:joint_framework}
\end{figure*}

\section{J\texorpdfstring{$\hat{\text{A}}$}{Lg}A-Net for Facial AU Detection and Face Alignment}

\subsection{Overview}

The architecture of our proposed J$\hat{\text{A}}$A-Net is shown in Fig.~\ref{fig:joint_framework}, which takes a color face with size $l\times l \times 3$ as input. It consists of four modules in different colors: hierarchical and multi-scale region learning, face alignment, global feature learning, and adaptive attention learning. Firstly, the hierarchical and multi-scale region learning is designed as the foundation of J$\hat{\text{A}}$A-Net, which extracts a multi-scale feature from local regions with different sizes. Secondly, the face alignment module is designed to estimate the locations of facial landmarks, which will be further utilized to predefine the initial attention map of each AU. The global feature learning is to capture the structure and texture information of the whole face. Finally, the adaptive attention learning (in red) is designed as the central part for AU detection with a multi-branch network, which refines the attention map of each AU adaptively so as to capture local AU features at different locations. The assembled local AU features are then integrated with the face alignment feature and the global feature for final AU detection. The three modules, face alignment, global feature learning, and adaptive attention learning, are optimized jointly, which share the layers of the hierarchical and multi-scale region learning.

\begin{figure}
\centering\includegraphics[width=\linewidth]{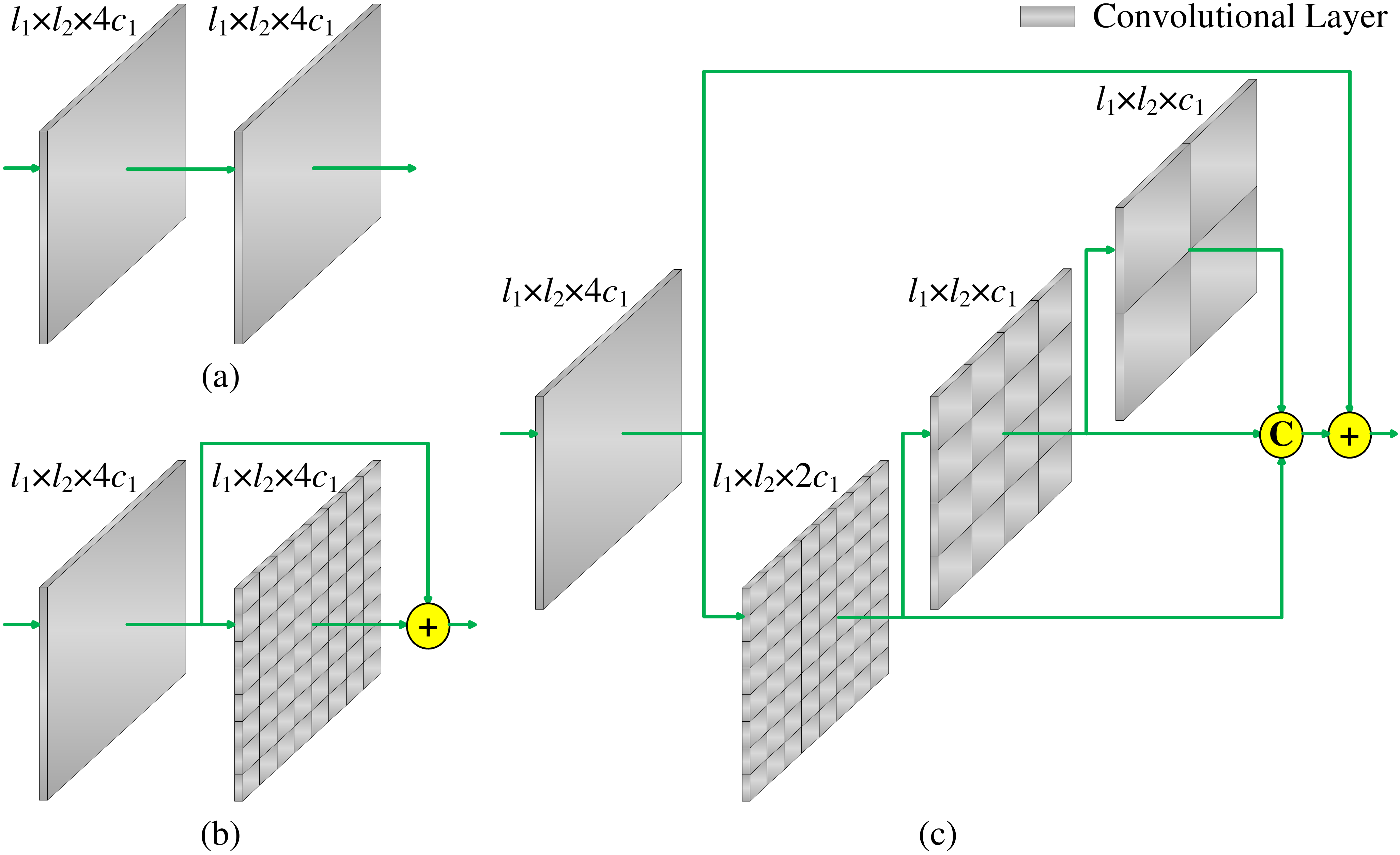}
\caption{The structures of different blocks for region learning. \textbf{(a)} A block of two plain convolutional layers $P(l_1,l_2,c_1)$. \textbf{(b)} A block of region layer $R(l_1,l_2,c_1)$. \textbf{(c)} A block of hierarchical and multi-scale region layer $R_{hm}(l_1,l_2,c_1)$. The expression of $l_1 \times l_2 \times c_1$ indicates that the height, width, and channel of a layer are $l_1$, $l_2$, and $c_1$, respectively. ``+'' denotes element-wise sum of feature maps.}
\label{fig:multi_region_layer}
\end{figure}

\subsection{Hierarchical and Multi-Scale Region Learning}

Considering AUs in different local facial regions have various structure and texture information, different local regions should be processed with different filters. However, a plain convolutional layer only uses a convolutional filter shared across the entire spatial domain. To extract more precise features of local regions, \cite{zhao2016deep} proposed a region layer $R(l_1,l_2,c_1)$ which contains a plain convolutional layer and a partitioned convolutional layer, as shown in Fig.~\ref{fig:multi_region_layer}(b). In the partitioned convolutional layer, each local patch has an independent convolutional filter. However, all the local patches have identical sizes, which limits the performance of the region layer to process various AUs with different sizes.

To address this issue, we propose a hierarchical and multi-scale region layer $R_{hm}(l_1,l_2,c_1)$ to learn features from multi-scale local regions. Fig.~\ref{fig:multi_region_layer}(c) illustrates its detailed structure. It consists of a plain convolutional layer and another three hierarchical partitioned convolutional layers. Specifically, the feature map of the plain convolutional layer is uniformly divided into $8 \times 8$ patches, each of which is processed with an independent convolutional filter by the first partitioned convolutional layer. In the same manner, the second and third partitioned convolutional layers apply independent convolutional filters on the uniformly divided $4 \times 4$ and $2 \times 2$ feature map patches of their previous layers, respectively. By concatenating the feature maps of the first, second, and third partitioned convolutional layers, we can extract a hierarchical and multi-scale feature map with the same number of channels $4c_1$ as the feature map of the plain convolutional layer. A residual structure~\citep{he2016deep} is then utilized to element-wise sum the two feature maps, so as to learn over-complete features and avoid the vanishing gradient problem. Different from the region layer~\citep{zhao2016deep}, our proposed hierarchical and multi-scale region layer uses multi-scale partitions, which are beneficial for covering all kinds of AUs in the ROIs with different sizes.

In our J$\hat{\text{A}}$A-Net, this hierarchical and multi-scale region learning module is composed by two blocks of $R_{hm}(l,l,c)$ and $R_{hm}(l/2,l/2,2c)$, each of which is followed by a max-pooling layer to reduce its feature map size. $c$ is a parameter with respect to the number of layer channels. The multi-scale feature with size $l/4\times l/4\times 8c$ output by this module is further fed into the rest three modules to facilitate both AU detection and face alignment.

\begin{figure*}
\centering\includegraphics[width=\linewidth]{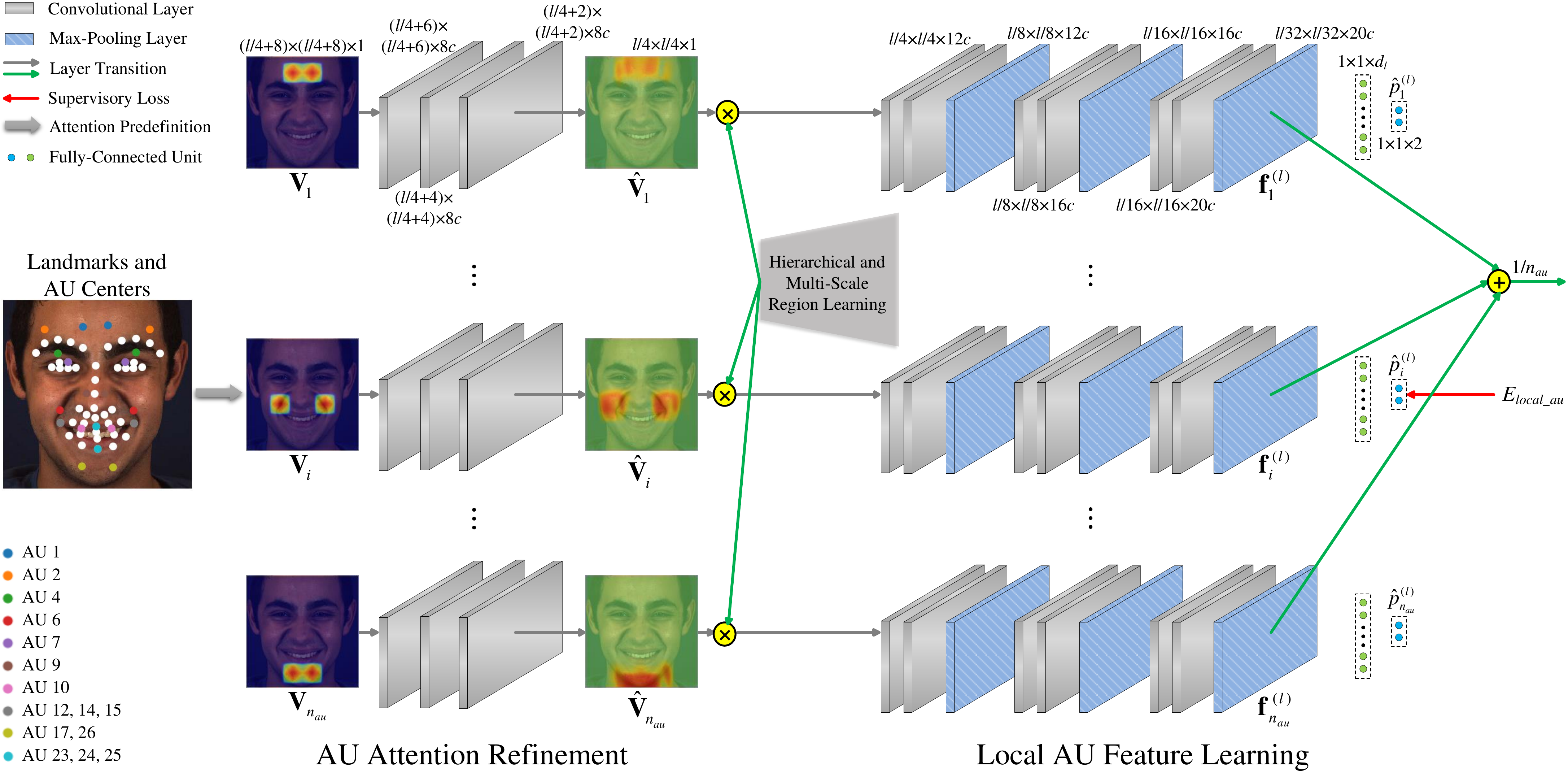}
\caption{The architecture of the proposed adaptive attention learning, which consists of two steps horizontally, i.e. the AU attention refinement step and the local AU feature learning step,  and multiple branches with the same structure vertically, one branch for one AU. The three shown attention maps correspond to AUs 1, 6, and 17, respectively. In the local AU feature learning step, each two plain convolutional layers have the same size. ``+'' followed by ``$1/n_{au}$'' denotes the element-wise average over all the local AU feature maps. Note that the cropping of attention maps in the earlier conference version~\citep{shao2018deep} is removed due to its redundancy.}
\label{fig:adaptive_attention}
\end{figure*}

\subsection{Face Alignment and Global Feature Learning}

The face alignment module includes three successive blocks of $P(l/4,l/4,3c)$, $P(l/8,l/8,4c)$, and $P(l/16,l/16,5c)$, each of which is followed by a max-pooling layer. \highlight{This module outputs a face alignment feature, which contains global facial shape information and local landmark information.} As shown in Fig.~\ref{fig:joint_framework}, the face alignment feature is fed into two fully-connected layers with the dimensions of $d$ and $2n_{align}$ respectively, where $n_{align}$ is the number of facial landmarks. We define the face alignment loss as
\begin{equation}\label{eq:Ealign}
E_{align} = \frac{1}{2d_o^2}\sum_{j=1}^{n_{align}} [(y_{2j-1}-\hat{y}_{2j-1})^2+(y_{2j}-\hat{y}_{2j})^2],
\end{equation}
where $y_{2j-1}$ and $y_{2j}$ denote the ground-truth $x$-coordinate and $y$-coordinate of the $j$-th facial landmark, $\hat{y}_{2j-1}$ and $\hat{y}_{2j}$ are the corresponding predicted results. $d_o$ is the ground-truth inter-ocular distance for normalization~\citep{shao2016learning,shao2019deep}.

The global feature learning module is utilized to capture global facial structure and texture information, which has the same structure as the face alignment module. As illustrated in Fig.~\ref{fig:joint_framework}, its output global feature and the face alignment feature are both used for the final AU detection, which can provide complementary useful information on top of the local AU features.

\subsection{Adaptive Attention Learning}

Fig.~\ref{fig:adaptive_attention} shows the architecture of the proposed adaptive attention learning. It consists of two steps: AU attention refinement and local AU feature learning, where the first step is to refine the predefined attention map of each AU with a separate branch and the second step is to learn and extract local AU features.

\begin{table}
\centering\caption{Definitions for the locations of AU centers, which are applicable to an aligned face with eye centers on the same horizontal line. ``Scale'' denotes the distance between two inner eye corners. Besides the definitions for $12$ AUs in~\cite{li2018eac}, we supplement the definitions for AUs 9, 25, and 26.}
\label{tab:au_definition}
\begin{tabular}{|*{3}{c|}}
\hline
AU &Description &Location\\\hline
1 &Inner brow raiser &1/2 scale above inner brow\\\hline
2 &Outer brow raiser &1/3 scale above outer brow\\\hline
4 &Brow lowerer &1/3 scale below brow center\\\hline
6 &Cheek raiser &1 scale below eye bottom\\\hline
7 &Lid tightener &Eye\\\hline
9 &Nose wrinkler &1/2 scale above nose bottom\\\hline
10 &Upper lip raiser &Upper lip center\\\hline
12 &Lip corner puller &Lip corner\\\hline
14 &Dimpler &Lip corner\\\hline
15 &Lip corner depressor &Lip corner\\\hline
17 &Chin raiser &1/2 scale below lip\\\hline
23 &Lip tightener &Lip center\\\hline
24 &Lip pressor &Lip center\\\hline
25 &Lips part &Lip center\\\hline
26 &Jaw drop &1/2 scale below lip\\\hline
\end{tabular}
\end{table}

\noindent\textbf{Predefinition of Attention Maps.}
In the AU attention refinement step, we first utilize the estimated facial landmarks by the face alignment module to predefine the locations of AU centers. The rules for defining the locations of AU centers are detailed in Table~\ref{tab:au_definition}. We also visualize the locations of AU centers and landmarks on the input image in Fig.~\ref{fig:adaptive_attention}, in which each AU has two centers due to the symmetry. The predefined attention map of each AU contains two sub-regions of interest (sub-ROIs) centered around the AU centers. Let the size of predefined attention maps be $l_{a_{pre}}\times l_{a_{pre}}\times 1$, we need to convert the locations of AU centers from the image scale to current map scale by multiplying $l_{a_{pre}}/l$ with both x- and y-coordinates of AU centers. The size of each sub-ROI is set to be $l_{a_{pre}}\zeta\times l_{a_{pre}}\zeta$, where $\zeta$ is the width ratio between the sub-ROI and the predefined attention map.

The predefined attention maps are set to only highlight in the two sub-ROIs, in which the attention weight of any point beyond the sub-ROIs is initialized to be $0$. Specifically, for the predefined attention map $\mathbf{V}_i$ of the $i$-th AU, if the $k$-th point is in a sub-ROI, its attention weight is initialized as
\begin{equation} \label{eq:vik}
v_{ik} = \max \{1 - \frac{d_{ik} \xi}{l_{a_{pre}}\zeta}, 0\}, \quad i=1, \cdots, n_{au},
\end{equation}
where $d_{ik}$ is the Manhattan distance of this point to the AU center, and $n_{au}$ is the number of AUs. $\xi \geq0$ is a coefficient to control the intensity of attention weights, in which the attention weights of all the sub-ROI points will become $1$ if $\xi=0$. Eq.~\eqref{eq:vik} essentially suggests that the attention weights are decaying when the sub-ROI points are moving away from the AU center. The maximization operation in Eq.~\eqref{eq:vik} is to ensure $v_{ik} \in [0,1]$. If a point belongs to the overlap of two sub-ROIs, it is set to be the maximum value of two computed attention weights associated with each sub-ROI.

\noindent\textbf{Refinement of Attention Maps.}
In each AU attention refinement branch, we employ three plain convolutional layers with $8c$ channels and a plain convolutional layer with one channel to process the predefined attention map, in which the last plain convolutional layer followed by a sigmoid function outputs the refined attention map $\hat{\mathbf{V}}_i$. The sigmoid function is to make each refined attention weight $\hat{v}_{ik}\in (0,1)$, in which the attention distributions of sub-ROIs and the remaining regions are both adaptively refined.

\begin{figure}
\centering\includegraphics[width=\linewidth]{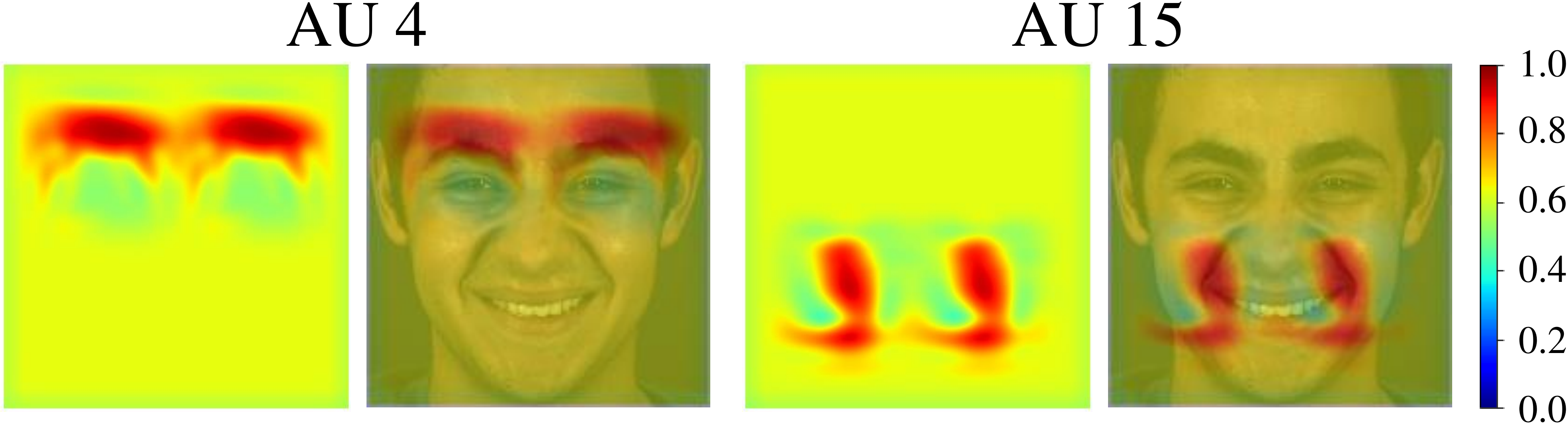}
\caption{\highlight{The effect of padding in plain convolutional layers of the AU attention refinement for two example AUs, in which the narrow regions along four boundary edges have distinct attention weights (zoom-in for best view). The colors from blue to red in the color bar denote attention weights from $0$ to $1$.}}
\label{fig:padding_effect}
\end{figure}

Considering that padding in plain convolutional layers could do harm to the refinement of edge regions in attention maps, as illustrated in Fig.~\ref{fig:padding_effect}, we set the padding of convolutional filters in all the four plain convolutional layers to be $0$. In this case, the sizes of their feature maps will be reduced after each convolution. As shown in Fig.~\ref{fig:adaptive_attention}, to match the size $l/4\times l/4\times 8c$ of the multi-scale feature output by the hierarchical and multi-scale region learning module, the size of the refined attention map should be $l/4\times l/4\times 1$. Therefore, we set $l_{a_{pre}}=l/4+8$ so that the feature map widths of the four plain convolutional layers become $l/4+6$, $l/4+4$, $l/4+2$, and $l/4$, respectively.

\noindent\textbf{Learning of Local AU Features.}
In the local AU feature learning step, the refined attention map of each AU is first element-wise multiplied with the multi-scale feature to obtain AU attention weighted feature. Then, each branch is performed with a network consisting of three max-pooling layers, each of which follows a block of two plain convolutional layers defined in Fig.~\ref{fig:multi_region_layer}(a). The last max-pooling layer outputs the learned local AU feature $\mathbf{f}_i^{(l)}$, and all the local AU features are assembled:
\begin{equation}
    \mathbf{f}^{(l)}=\frac{1}{n_{au}}\sum_{i=1}^{n_{au}}\mathbf{f}_i^{(l)}.
\end{equation}
The assembled local features $\mathbf{f}^{(l)}$ capture precise local texture information with respect to AUs, which will then contribute to the final AU detection. Note that the adaptive attention refinement requires the supervision of AU detection. To supervise the learning of each AU attention map, we apply a local AU detection loss $E_{local\_au}$ to each AU branch. The details of $E_{local\_au}$ will be elaborated in Sec.~\ref{ssec:au_det}.

\subsection{Facial AU Detection}
\label{ssec:au_det}

\noindent\textbf{AU Detection Using Integrated Information.}
As illustrated in Fig.~\ref{fig:joint_framework}, \highlight{the face alignment feature with global facial shape information and local landmark information, the global feature with global facial structure and texture information, and the assembled local features with local texture information} are concatenated together and fed into two fully-connected layers with the dimensions of $d$ and $2n_{au}$, respectively. \highlight{In this way, the various useful information is integrated together for facial AU detection.} Finally, a softmax function is applied across each two units of the last $2n_{au}$-dimensional fully-connected layer to obtain the predicted occurrence probability $\hat{p}_{i}$ of each AU.

Facial AU detection can be regarded as a multi-label binary classification problem with the following weighted multi-label cross entropy loss:
\begin{equation}\label{eq:Ecross}
E_{cross} = -\sum_{i=1}^{n_{au}} w_i [p_{i} \log \hat{p}_{i} + (1-p_{i}) \log (1-\hat{p}_{i})],
\end{equation}
where $p_{i}$ denotes the ground-truth occurrence probability of the $i$-th AU, which is $1$ if occurrence and $0$ otherwise. The weight $w_i$ introduced in Eq.~\eqref{eq:Ecross} is to alleviate the data imbalance problem. For most facial AU detection benchmarks, the occurrence rates of AUs are imbalanced~\citep{TAC2017Pantic}. Since AUs are not mutually independent, imbalanced training data has a bad influence on this multi-label learning task. Particularly, we set $w_i = (1/r_i)/\sum_{u=1}^{n_{au}}(1/r_u)$, where $r_i$ is the occurrence rate of the $i$-th AU in the training set.

In many cases, some AUs appear rarely in training samples, for which the cross entropy loss in Eq.~\eqref{eq:Ecross} often makes the AU prediction strongly bias towards non-occurrence. To address this, we exploit precision and recall which are both relevant to the true positive. Since F1-score: $F1=2PR/(P+R)$ considers both precision $P$ and recall $R$, we introduce a weighted multi-label Dice coefficient (F1-score) loss~\citep{milletari2016v}:
\begin{equation}
\label{eq:Edice}
E_{dice} = \sum_{i=1}^{n_{au}} w_i (1-\frac{2 p_{i} \hat{p}_{i} + \epsilon}{p_{i}^2 + \hat{p}_{i}^2 + \epsilon}),
\end{equation}
where $\epsilon$ is a smooth term. F1-score is known as the most popular evaluation metric for facial AU detection. The use of Eq.~\eqref{eq:Edice} keeps the consistency between the learning process and the evaluation metric. By combining Eqs.~\eqref{eq:Ecross} and~\eqref{eq:Edice}, we can obtain the overall AU detection loss:
\begin{equation}\label{eq:Eau}
E_{all\_au} = E_{cross} + E_{dice},
\end{equation}
where the occurrence probability $\hat{p}_i$ of each AU is predicted based on the integrated information of all the AUs.

\noindent\textbf{Local AU Detection for Adaptive Attention Learning.}
As illustrated in Fig.~\ref{fig:adaptive_attention}, the local AU detection loss $E_{local\_au}$ is used to supervise the adaptive refinement of each AU attention map. In particular, the local feature $\mathbf{f}_i^{(l)}$ of the $i$-th AU is followed by two fully-connected layers with the dimensions of $d_l$ and $2$ respectively, in which a softmax function is applied to the last layer to predict a temporary occurrence probability $\hat{p}_{i}^{(l)}$. \highlight{Similar to Eq.~\eqref{eq:Eau}, $E_{local\_au}$ is defined as}
\begin{equation}\label{eq:Elocal_au}
\begin{aligned}
E_{local\_au} = -&\sum_{i=1}^{n_{au}} w_i [p_{i} \log \hat{p}_{i}^{(l)} + (1-p_{i}) \log (1-\hat{p}_{i}^{(l)})]+\\
&\sum_{i=1}^{n_{au}} w_i (1-\frac{2 p_{i} \hat{p}_{i}^{(l)} + \epsilon}{p_{i}^2 + (\hat{p}_{i}^{(l)})^2 + \epsilon}),
\end{aligned}
\end{equation}
where $\hat{p}_{i}^{(l)}$ is predicted only based on the local information of the $i$-th AU. The use of local information is beneficial for avoiding the influence of irrelevant AUs on the attention refinement. Due to the same reason, the assembled local features $\mathbf{f}^{(l)}$ are set to discard the back-propagated gradients from $E_{all\_au}$ which uses the integrated information of all the AUs. In this case, $E_{local\_au}$ is also responsible for supervising the learning of each local feature $\mathbf{f}^{(l)}_i$.

After introducing each loss term, we can define the overall loss of our J$\hat{\text{A}}$A-Net as
\begin{equation}
E= (E_{all\_au}+E_{local\_au}) + \lambda_{align} E_{align},
\end{equation}
where $\lambda_{align}$ is a trade-off parameter. Our framework is trainable end-to-end, in which the hierarchical and multi-scale region learning, face alignment, global feature learning, and adaptive attention learning are trained simultaneously. The joint training of facial AU detection and face alignment contributes to each other due to the close relationship between the two tasks.

\begin{table*}
\centering\caption{F1-frame and accuracy results for $12$ AUs on BP4D~\citep{zhang2014bp4d}. The results of LSVM~\citep{fan2008liblinear} and JPML~\citep{zhao2016joint} are from~\cite{zhao2016deep}, those of CPM~\citep{zeng2015confidence} are from~\cite{chu2017learning}, and those of all the remaining methods are directly taken from their original papers.}
\label{tab:comp_f1_acc_bp4d}
\begin{tabular}{|*{15}{c|}}
\hline
\multirow{2}*{AU} &\multicolumn{10}{c|}{F1-Frame}  &\multicolumn{4}{c|}{Accuracy} \\
\cline{2-15} &\rotatebox[origin=c]{90}{\tabincell{c}{LSVM\\\citep{fan2008liblinear}}} &\rotatebox[origin=c]{90}{\tabincell{c}{JPML\\\citep{zhao2016joint}}} &\rotatebox[origin=c]{90}{\tabincell{c}{DRML\\\citep{zhao2016deep}}} &\rotatebox[origin=c]{90}{\tabincell{c}{CPM\\\citep{zeng2015confidence}}} &\rotatebox[origin=c]{90}{\tabincell{c}{EAC-Net\\\citep{li2018eac}}} &\rotatebox[origin=c]{90}{\tabincell{c}{DSIN\\\citep{corneanu2018deep}}} &\rotatebox[origin=c]{90}{\tabincell{c}{CMS\\\citep{sankaran2019representation}}} &\rotatebox[origin=c]{90}{\tabincell{c}{LP-Net\\\citep{niu2019local}}} &\rotatebox[origin=c]{90}{\tabincell{c}{ARL\\\citep{shao2019facial}}} &\rotatebox[origin=c]{90}{\textbf{J$\hat{\text{A}}$A-Net}} &\rotatebox[origin=c]{90}{\tabincell{c}{EAC-Net\\\citep{li2018eac}}} &\rotatebox[origin=c]{90}{\tabincell{c}{CMS\\\citep{sankaran2019representation}}} &\rotatebox[origin=c]{90}{\tabincell{c}{ARL\\\citep{shao2019facial}}} &\rotatebox[origin=c]{90}{\textbf{J$\hat{\text{A}}$A-Net}}\\\hline
1 &23.2 &32.6 &36.4 &43.4 &39.0 &51.7 &49.1 &43.4 &45.8 &\textbf{53.8} &68.9 &55.9 &73.9 &\textbf{75.2}\\
2 &22.8 &25.6 &41.8 &40.7 &35.2 &40.4 &44.1 &38.0 &39.8 &\textbf{47.8} &73.9 &67.7 &76.7 &\textbf{80.2}\\
4 &23.1 &37.4 &43.0 &43.3 &48.6 &56.0 &50.3 &54.2 &55.1 &\textbf{58.2} &78.1 &71.5 &80.9 &\textbf{82.9}\\
6 &27.2 &42.3 &55.0 &59.2 &76.1 &76.1 &\textbf{79.2} &77.1 &75.7 &78.5 &78.5 &\textbf{81.3} &78.2 &79.8\\
7 &47.1 &50.5 &67.0 &61.3 &72.9 &73.5 &74.7 &76.7 &\textbf{77.2} &75.8 &69.0 &71.9 &\textbf{74.4} &72.3\\
10 &77.2 &72.2 &66.3 &62.1 &81.9 &79.9 &80.9 &\textbf{83.8} &82.3 &82.7 &77.6 &77.3 &\textbf{79.1} &78.2\\
12 &63.7 &74.1 &65.8 &68.5 &86.2 &85.4 &\textbf{88.3} &87.2 &86.6 &88.2 &84.6 &\textbf{87.4} &85.5 &86.6\\
14 &64.3 &\textbf{65.7} &54.1 &52.5 &58.8 &62.7 &63.9 &63.3 &58.8 &63.7 &60.6 &57.4 &62.8 &\textbf{65.1}\\
15 &18.4 &38.1 &33.2 &36.7 &37.5 &37.3 &44.4 &45.3 &\textbf{47.6} &43.3 &78.1 &71.6 &\textbf{84.7} &81.0\\
17 &33.0 &40.0 &48.0 &54.3 &59.1 &\textbf{62.9} &60.3 &60.5 &62.1 &61.8 &70.6 &73.7 &\textbf{74.1} &72.8\\
23 &19.4 &30.4 &31.7 &39.5 &35.9 &38.8 &41.4 &\textbf{48.1} &47.4 &45.6 &81.0 &74.6 &\textbf{82.9} &\textbf{82.9}\\
24 &20.7 &42.3 &30.0 &37.8 &35.8 &41.6 &51.2 &54.2 &\textbf{55.4} &49.9 &82.4 &84.1 &85.7 &\textbf{86.3}\\
\hline
Avg &35.3 &45.9 &48.3 &50.0 &55.9 &58.9 &60.6 &61.0 &61.1 &\textbf{62.4} &75.2 &72.9 &78.2 &\textbf{78.6}\\
\hline
\end{tabular}
\end{table*}

\section{Experiments}

\subsection{Datasets and Settings}

\subsubsection{Datasets}

Our J$\hat{\text{A}}$A-Net is evaluated on four widely used datasets for AU detection, i.e. BP4D~\citep{zhang2014bp4d}, DISFA~\citep{mavadati2013disfa}, GFT~\citep{girard2017sayette}, and BP4D+~\citep{zhang2016multimodal}, in which both AU and landmark labels are provided.
\begin{itemize}
    \item\textbf{BP4D} contains $41$ subjects with $23$ females and $18$ males, each of which is involved in $8$ sessions. There are $328$ videos including about $140,000$ frames with AU labels of occurrence or non-occurrence. Each frame is also annotated with $49$ landmarks detected by SDM~\citep{xiong2013supervised}. Similar to the settings of~\cite{zhao2016deep,li2018eac}, $12$ AUs (1, 2, 4, 6, 7, 10, 12, 14, 15, 17, 23, and 24) are evaluated using subject exclusive 3-fold cross-validation, where two folds are used for training and the remaining one is used for testing.
    \item\textbf{DISFA} consists of $27$ videos recorded from $12$ women and $15$ men, each of which has $4,845$ frames. Each frame is annotated with AU intensities on a six-point ordinal scale from $0$ to $5$, as well as $66$ landmarks detected by AAM~\citep{cootes2001active}. To be consistent with BP4D, we use $49$ landmarks, a subset of the $66$ landmarks. It has a serious data imbalance problem, in which most AUs have very low occurrence rates while only a few other AUs have higher occurrence rates. According to the setting in~\cite{zhao2016deep,li2018eac}, AU intensities equal or greater than $2$ are considered as occurrence, while others are treated as non-occurrence. Subject exclusive 3-fold cross-validation is also conducted with evaluations on $8$ AUs (1, 2, 4, 6, 9, 12, 25, and 26).
    \item\textbf{GFT} includes $96$ subjects from $32$ three-subject groups with unscripted social communication. Each subject is captured using a video with annotations of $10$ AUs (1, 2, 4, 6, 10, 12, 14, 15, 23, and 24), as well as $49$ landmarks detected by ZFace~\citep{jeni2017dense}. The captured frames exhibit moderate out-of-plane poses, resulting in a more challenging AU detection. Following the original training/testing partitions in~\cite{girard2017sayette}, we employ $78$ subjects with about $108,000$ frames for training, and $18$ subjects with about $24,600$ frames for testing.
    \item\textbf{BP4D+} consists of $140$ subjects with $82$ females and $58$ males, each of which is involved in $10$ sessions. \highlight{Compared with BP4D~\citep{zhang2014bp4d} dataset, it has similar style but larger scale and variability.} AU annotations are provided for $4$ sessions with totally $197,875$ frames, in which each frame is also detected with $49$ landmarks by ZFace~\citep{jeni2017dense}. \highlight{To evaluate the performance on large-scale testing data,} similar to~\cite{shao2019facial}, we train on the entire BP4D dataset ($41$ subjects with $12$ AUs) and test on all the BP4D+ images.
\end{itemize}

\begin{table*}
\centering\caption{F1-frame and accuracy results for $8$ AUs on DISFA~\citep{mavadati2013disfa}. \highlight{The results of LSVM~\citep{fan2008liblinear} and APL~\citep{zhong2015learning} are from~\cite{zhao2016deep}, and those of all the remaining methods are directly taken from their original papers.}}
\label{tab:comp_f1_acc_disfa}
\begin{tabular}{|*{14}{c|}}
\hline
\multirow{2}*{AU} &\multicolumn{9}{c|}{F1-Frame} &\multicolumn{4}{c|}{Accuracy}\\
\cline{2-14}&\rotatebox[origin=c]{90}{\tabincell{c}{LSVM\\\citep{fan2008liblinear}}}
&\rotatebox[origin=c]{90}{\tabincell{c}{APL\\\citep{zhong2015learning}}}
&\rotatebox[origin=c]{90}{\tabincell{c}{DRML\\\citep{zhao2016deep}}}
&\rotatebox[origin=c]{90}{\tabincell{c}{EAC-Net\\\citep{li2018eac}}}
&\rotatebox[origin=c]{90}{\tabincell{c}{DSIN\\\citep{corneanu2018deep}}}
&\rotatebox[origin=c]{90}{\tabincell{c}{CMS\\\citep{sankaran2019representation}}}
&\rotatebox[origin=c]{90}{\tabincell{c}{LP-Net\\\citep{niu2019local}}}
&\rotatebox[origin=c]{90}{\tabincell{c}{ARL\\\citep{shao2019facial}}}
&\rotatebox[origin=c]{90}{\textbf{J$\hat{\text{A}}$A-Net}}
&\rotatebox[origin=c]{90}{\tabincell{c}{EAC-Net\\\citep{li2018eac}}}
&\rotatebox[origin=c]{90}{\tabincell{c}{CMS\\\citep{sankaran2019representation}}}
&\rotatebox[origin=c]{90}{\tabincell{c}{ARL\\\citep{shao2019facial}}}
&\rotatebox[origin=c]{90}{\textbf{J$\hat{\text{A}}$A-Net}}\\\hline
1 &10.8 &11.4 &17.3 &41.5 &42.4 &40.2 &29.9 &43.9 &\textbf{62.4} &85.6 &91.6 &92.1 &\textbf{97.0}\\
2 &10.0 &12.0 &17.7 &26.4 &39.0 &44.3 &24.7 &42.1 &\textbf{60.7} &84.9 &94.7 &92.7 &\textbf{97.3}\\
4 &21.8 &30.1 &37.4 &66.4 &68.4 &53.2 &\textbf{72.7} &63.6 &67.1 &79.1 &79.9 &\textbf{88.5} &88.0\\
6 &15.7 &12.4 &29.0 &50.7 &28.6 &\textbf{57.1} &46.8 &41.8 &41.1 &69.1 &82.6 &91.6 &\textbf{92.1}\\
9 &11.5 &10.1 &10.7 &\textbf{80.5} &46.8 &50.3 &49.6 &40.0 &45.1 &88.1 &95.2 &\textbf{95.9} &95.6\\
12 &70.4 &65.9 &37.7 &\textbf{89.3} &70.8 &73.5 &72.9 &76.2 &73.5 &90.0 &87.8 &\textbf{93.9} &92.3\\
25 &12.0 &21.4 &38.5 &88.9 &90.4 &81.1 &93.8 &\textbf{95.2} &90.9 &80.5 &86.3 &\textbf{97.3} &94.9\\
26 &22.1 &26.9 &20.1 &15.6 &42.2 &59.7 &65.0 &66.8 &\textbf{67.4} &64.8 &80.7 &94.3 &\textbf{94.8}\\
\hline
Avg &21.8 &23.8 &26.7 &48.5 &53.6 &57.4 &56.9 &58.7 &\textbf{63.5} &80.6 &87.3 &93.3 &\textbf{94.0}\\
\hline
\end{tabular}
\end{table*}

\begin{table*}
\centering\caption{F1-frame and accuracy results for $10$ AUs on GFT~\citep{girard2017sayette}. The results of LSVM~\citep{fan2008liblinear} and AlexNet~\citep{krizhevsky2012imagenet} are from~\cite{girard2017sayette}, and EAC-Net~\citep{li2018eac} is implemented on GFT using its released code.}
\label{tab:gft_AUocc}
\begin{tabular}{|*{13}{c|}}
\hline
\multicolumn{2}{|c|}{AU}&1&2&4&6&10&12&14&15&23&24&Avg\\
\hline
\multirow{4}*{F1-Frame}&LSVM~\citep{fan2008liblinear} &38 &32 &13 &67 &64
&78 &15 &29 &49 &44 &42.9\\
&AlexNet~\citep{krizhevsky2012imagenet} &44 &46 &2 &73 &72 &82 &5 &19 &43 &42 &42.8\\
&\highlight{EAC-Net~\citep{li2018eac}} &15.5 &\textbf{56.6} &0.1 &\textbf{81.0} &76.1 &84.0 &0.1 &38.5 &57.8 &\textbf{51.2} &46.1\\
&\highlight{TCAE~\citep{li2019self-supervised}} &43.9 &49.5 &6.3
&71.0 &76.2 &79.5 &10.7 &28.5 &34.5 &41.7 &44.2\\
&ARL~\citep{shao2019facial} &\textbf{51.9} &45.9 &13.7 &79.2 &75.5 &82.8 &0.1 &\textbf{44.9} &\textbf{59.2} &47.5 &50.1\\
&\highlight{\cite{ertugrul2020crossing}} &43.7 &44.9 &\textbf{19.8} &74.6 &\textbf{76.5} &79.8 &\textbf{50.0} &33.9 &16.8 &12.9 &45.3\\
&\textbf{J$\hat{\text{A}}$A-Net} &46.5 &49.3 &19.2 &79.0 &75.0 &\textbf{84.8} &44.1 &33.5 &54.9 &50.7 &\textbf{53.7}\\
\hline
\multirow{2}*{Accuracy}&\highlight{EAC-Net~\citep{li2018eac}} &94.8 &\textbf{89.0} &95.6 &\textbf{89.6} &\textbf{88.5} &90.4 &96.0 &91.3 &80.2 &84.8 &90.0\\
&ARL~\citep{shao2019facial}&\textbf{96.6} &82.4 &96.1 &89.0 &87.5 &90.3 &97.0 &91.1 &\textbf{80.4} &83.6 &89.4\\
&\textbf{J$\hat{\text{A}}$A-Net}&96.2 &87.4 &\textbf{96.3} &89.4 &\textbf{88.5} &\textbf{91.8} &\textbf{97.2} &\textbf{91.4} &79.8 &\textbf{86.7} &\textbf{90.5}\\
\hline
\end{tabular}
\end{table*}

\subsubsection{Implementation Details}

For each face image, we perform similarity transformation including \highlight{in-plane rotation}, uniform scaling, and translation to obtain an aligned $200 \times 200 \times 3$ color face. This transformation is shape-preserving and brings no change to the expression. In order to enhance the diversity of training data, aligned faces are further randomly cropped into $176\times 176$ and horizontally flipped. In our J$\hat{\text{A}}$A-Net, each convolutional layer uses $3\times 3$ convolutional filters with a stride $1$ and a padding $1$ except for the AU attention refinement layers using a padding $0$, and each max-pooling layer processes $2\times 2$ spatial fields with a stride $2$ and a padding $0$. Besides, each convolutional layer is operated with Batch Normalization (BN)~\citep{ioffe2015batch} and Rectified Linear Unit (ReLU)~\citep{nair2010rectified}. Following the settings in~\cite{zhao2016deep,li2018eac}, J$\hat{\text{A}}$A-Net is initialized with the parameters of the well-trained model for BP4D when trained on DISFA.

Our J$\hat{\text{A}}$A-Net is implemented using PyTorch~\citep{paszke2019pytorch} with the stochastic gradient descent (SGD) solver, a Nesterov momentum~\citep{sutskever2013importance} of $0.9$, and a weight decay of $0.0005$. We train J$\hat{\text{A}}$A-Net for up to $12$ epochs with initial learning rate of $0.01$, in which the learning rate is multiplied by a factor of $0.3$ at every $2$ epochs. The parameters with respect to the structure of J$\hat{\text{A}}$A-Net are chosen as $l= 176$, $c=8$, $d=512$, and $d_l=64$. Besides, $n_{align} = 49$, $n_{au}$ is $12$, $8$, $10$, and $12$ for BP4D, DISFA, GFT, and BP4D+, respectively, $\epsilon=1$, and $\zeta=0.14$ and $\xi=0.56$ are used to generate an approximate Gaussian attention distribution for each AU sub-ROI. \highlight{To set an appropriate value for the trade-off parameter $\lambda_{align}$, we select multiple small sets from training data as validation sets, and perform cross-validation on these sets. In each validation experiment, we train J$\hat{\text{A}}$A-Net on the training data excluding the current small validation set. $\lambda_{align}$ is set to $0.5$ for the overall best performance on the validation sets.}

\subsubsection{Evaluation Metrics}

The evaluation metrics for the two tasks are chosen as follows:
\begin{itemize}
    \item\textbf{Facial AU Detection:} Following the previous methods of~\cite{zhao2016deep,li2018eac}, the frame-based F1-score (F1-frame, $\%$) is reported. To conduct a more comprehensive comparison, we also evaluate the performance with accuracy ($\%$). In addition, we compute the average results over all AUs (Avg). In the following sections, we omit $\%$ in all the results for simplicity.
    \item\textbf{Face Alignment:} We report the mean error normalized by inter-ocular distance, and treat the mean error larger than $10\%$ as a failure. In other words, we evaluate face alignment methods on the two popular metrics~\citep{kazemi2014one,zhang2016learning,shao2019deep}: mean error ($\%$) and failure rate ($\%$), where \% is also omitted in the results.
\end{itemize}

\subsection{Comparison with State-of-the-Art Methods}
\label{ssec:comp}

We compare our method J$\hat{\text{A}}$A-Net against state-of-the-art single-frame based facial AU detection methods under the same evaluation setting. These methods include LSVM~\citep{fan2008liblinear}, JPML~\citep{zhao2016joint}, APL~\citep{zhong2015learning}, AlexNet~\citep{krizhevsky2012imagenet}, DRML~\citep{zhao2016deep}, CPM~\citep{zeng2015confidence}, EAC-Net~\citep{li2018eac}, DSIN~\citep{corneanu2018deep}, CMS~\citep{sankaran2019representation}, LP-Net~\citep{niu2019local}, \highlight{TCAE~\citep{li2019self-supervised}}, ARL~\citep{shao2019facial}, and \highlight{\cite{ertugrul2020crossing}}. Note that a few related works like CNN+LSTM~\citep{chu2017learning} and R-T1~\citep{li2017action} are not compared due to their inputs of a sequence of frames instead of a single frame.

\noindent\textbf{Evaluation on BP4D.}
Table~\ref{tab:comp_f1_acc_bp4d} reports the F1-frame and accuracy results of different methods on BP4D. It can be seen that our J$\hat{\text{A}}$A-Net overall outperforms all previous works in terms of both F1-frame and accuracy metrics. J$\hat{\text{A}}$A-Net is superior to all the conventional methods including LSVM, JPML, and CPM, which demonstrates the strength of deep learning based methods. Compared to the latest LP-Net and ARL methods, J$\hat{\text{A}}$A-Net still achieves higher average F1-frame and average accuracy. Note that CMS shows good F1-frame results but poor accuracy results. In contrast, our method obtains high accuracy without sacrificing F1-frame, which is attributed to the integration of the cross entropy loss and the Dice coefficient loss in Eq.~\eqref{eq:Eau}.

\noindent\textbf{Evaluation on DISFA.}
Experimental results on DISFA are shown in Table~\ref{tab:comp_f1_acc_disfa}, from which it can be observed that our J$\hat{\text{A}}$A-Net outperforms all the state-of-the-art works with even more significant improvements. Specifically, J$\hat{\text{A}}$A-Net increases the average F1-frame and average accuracy by large margins of $4.8$ and $0.7$ over ARL, respectively. Due to the serious data imbalance issue in DISFA, performances of different AUs fluctuate severely in most of the previous methods. For instance, \highlight{the F1-frame of AU 12} is far higher than that of other AUs for LSVM and APL. Although EAC-Net processes the imbalance problem explicitly, its detection result for AU 26 is much worse than other AUs. \highlight{In contrast, our method introduces the Dice coefficient loss to suppress the prediction bias, and uses $w_i$ to give larger importance for AUs with lower occurrence rates in Eq.~\eqref{eq:Eau}, which contribute to the better results.}

\begin{table*}
\centering\caption{F1-frame and accuracy results for $12$ AUs on large-scale BP4D+~\citep{zhang2016multimodal}. EAC-Net~\citep{li2018eac} is implemented on BP4D+ using its released code.}
\label{tab:bp4d_plus_AUocc}
\begin{tabular}{|*{15}{c|}}
\hline
\multicolumn{2}{|c|}{AU}&1&2&4&6&7&10&12&14&15&17&23&24&Avg\\
\hline
\multirow{2}*{F1-Frame}&\highlight{EAC-Net~\citep{li2018eac}}&38.0 &\textbf{37.5} &\textbf{32.6} &82.0 &83.4 &87.1 &85.1 &62.1 &44.5 &43.6 &45.0 &32.8 &56.1\\
&ARL~\citep{shao2019facial}&29.9 &33.1 &27.1 &81.5 &83.0 &84.8 &86.2 &59.7 &\textbf{44.6} &43.7 &48.8 &32.3 &54.6\\
&\textbf{J$\hat{\text{A}}$A-Net}&\textbf{39.7} &35.6 &30.7 &\textbf{82.4} &\textbf{84.7} &\textbf{88.8} &\textbf{87.0} &\textbf{62.2} &38.9 &\textbf{46.4} &\textbf{48.9} &\textbf{36.0} &\textbf {56.8}\\
\hline
\multirow{2}*{Accuracy}&\highlight{EAC-Net~\citep{li2018eac}}&78.5 &\textbf{85.1} &\textbf{88.3} &\textbf{81.6} &\textbf{80.0} &84.7 &82.7 &\textbf{61.5} &88.6 &75.4 &\textbf{84.2} &91.5 &81.9\\
&ARL~\citep{shao2019facial}&67.2 &82.8 &84.4 &80.3 &77.8 &80.7 &82.9 &59.1 &88.0 &75.1 &83.9 &\textbf{93.2} &79.6\\
&\textbf{J$\hat{\text{A}}$A-Net}&\textbf{79.3} &\textbf{85.1} &85.1 &80.8 &79.1 &\textbf{85.0} &\textbf{83.3} &61.3 &\textbf{89.8} &\textbf{82.1} &83.3 &90.7 &\textbf{82.1}\\
\hline
\end{tabular}
\end{table*}

\begin{table*}
\centering\caption{The structures of different variants of J$\hat{\text{A}}$A-Net. \textbf{R}: region layer~\citep{zhao2016deep}. \textbf{HMR}: hierarchical and multi-scale region layer. \textbf{GF}: global feature learning. \textbf{C}: multi-label cross entropy loss. \textbf{D}: multi-label Dice coefficient loss. \textbf{W}: $w_i$ for weighting the loss of each AU. \textbf{FA}: face alignment module. \textbf{IF}: integrating the face alignment feature, global feature, and assembled local features if they are available. \textbf{LF}: local AU feature learning with input attention maps. \textbf{AR}: AU attention refinement. \highlight{\textbf{BE}: back-propagation enhancement~\citep{shao2018deep}. $E_r$: refinement constraint~\citep{shao2018deep}.} ``w/o'' is the abbreviation of ``without''.}
\label{tab:variant_JAA}
\begin{tabular}{|*{14}{c|}}
\hline
Method&R&HMR&GF&C&D&W&FA&IF&LF&AR&BE&$E_{local\_au}$&$E_r$\\
\hline
R-Net &$\surd$ & &$\surd$ &$\surd$ & & & & & & & & &\\
H-Net & &$\surd$ &$\surd$ &$\surd$ & & & & & & & & &\\
HD-Net & &$\surd$ &$\surd$ &$\surd$ &$\surd$ & & & & & & & &\\
HDW-Net & &$\surd$ &$\surd$ &$\surd$ &$\surd$ &$\surd$ & & & & & & &\\
\hline
J-Net w/o IF& &$\surd$ &$\surd$ &$\surd$ &$\surd$ &$\surd$ &$\surd$ & & & & & &\\
J-Net& &$\surd$ &$\surd$ &$\surd$ &$\surd$ &$\surd$ &$\surd$ &$\surd$ & & & & &\\
JA-Net w/o GF& &$\surd$ & &$\surd$ &$\surd$ &$\surd$ &$\surd$ &$\surd$ &$\surd$ & & & &\\
JA-Net& &$\surd$ &$\surd$ &$\surd$ &$\surd$ &$\surd$ &$\surd$ &$\surd$ &$\surd$ & & & &\\
\hline
J$\hat{\text{A}}$A-Net w/o $E_{local\_au}$ & &$\surd$ &$\surd$ &$\surd$ &$\surd$ &$\surd$ &$\surd$ &$\surd$ &$\surd$ &$\surd$ & & &\\
\highlight{J$\hat{\text{A}}$A-Net w/o $E_{local\_au}$+BE} & &$\surd$ &$\surd$ &$\surd$ &$\surd$ &$\surd$ &$\surd$ &$\surd$ &$\surd$ &$\surd$ &$\surd$ & &\\
\highlight{J$\hat{\text{A}}$A-Net w/o $E_{local\_au}$+BE+$E_r$} & &$\surd$ &$\surd$ &$\surd$ &$\surd$ &$\surd$ &$\surd$ &$\surd$ &$\surd$ &$\surd$ &$\surd$ & &$\surd$\\
\textbf{J$\hat{\text{A}}$A-Net}& &$\surd$ &$\surd$ &$\surd$ &$\surd$ &$\surd$ &$\surd$ &$\surd$ &$\surd$ &$\surd$ & &$\surd$ &\\
\hline
\end{tabular}
\end{table*}

\noindent\textbf{Evaluation on GFT.} Table~\ref{tab:gft_AUocc} shows the F1-frame and accuracy results of our J$\hat{\text{A}}$A-Net and previous works on GFT. \highlight{Compared to the recently proposed EAC-Net, TCAE, ARL and \cite{ertugrul2020crossing},} J$\hat{\text{A}}$A-Net achieves better performance with average F1-frame and average accuracy of $53.7$ and $90.5$, respectively. We notice that the numerical results of our J$\hat{\text{A}}$A-Net on GFT are lower than those on BP4D and DISFA. \highlight{This is because BP4D and DISFA images are frontal or near-frontal faces while GFT images are more challenging with out-of-plane poses.} We will further evaluate our method under non-frontal poses in Sec.~\ref{ssec:non_frontal}.

\noindent\textbf{Evaluation on BP4D+.}
\highlight{Considering the scale and variability of testing data also influence the evaluation performance,} we train J$\hat{\text{A}}$A-Net on all the BP4D images and test on the entire BP4D+ dataset. \highlight{Table~\ref{tab:bp4d_plus_AUocc} presents the results of EAC-Net, ARL, and J$\hat{\text{A}}$A-Net on BP4D+. We can see that our J$\hat{\text{A}}$A-Net has higher average F1-frame and average accuracy than other works. When the scale and variability of testing data are significantly increased, J$\hat{\text{A}}$A-Net still achieves the overall best performance.}

\begin{table*}
\centering\caption{F1-frame and accuracy results for $12$ AUs of different variants of our J$\hat{\text{A}}$A-Net on BP4D.}
\label{tab:ablation_bp4d}
\begin{tabular}{|*{15}{c|}}
\hline
\multicolumn{2}{|c|}{AU}&1&2&4&6&7&10&12&14&15&17&23&24&Avg\\
\hline
\multirow{12}*{F1-Frame}&R-Net&39.9 &37.6 &48.2 &74.5 &71.3 &81.4 &84.7 &58.6 &28.2 &59.1 &35.9 &39.6 &54.9\\
&H-Net&43.9 &46.4 &51.4 &74.3 &70.0 &79.5 &85.1 &55.9 &28.9 &58.2 &35.4 &40.1 &55.8\\
&HD-Net&39.7 &41.5 &51.4 &75.4 &70.8 &\textbf{82.9} &85.3 &58.9 &33.3 &59.9 &40.7 &39.3 &56.6\\
&HDW-Net&43.4 &41.5 &51.3 &75.5 &72.3 &82.8 &85.9 &60.2 &34.0 &62.0 &38.2 &41.5 &57.4\\
&J-Net w/o IF&42.4 &40.2 &54.4 &76.0 &70.3 &81.4 &86.5 &59.2 &39.6 &61.1 &42.2 &42.3 &58.0\\
&J-Net&46.4 &44.2 &48.9 &73.9 &71.1 &78.5 &84.7 &60.2 &43.5 &60.0 &45.4 &49.0 &58.8\\
&JA-Net w/o GF&47.1 &42.8 &51.5 &75.0 &74.2 &78.7 &85.8 &58.1 &45.6 &58.2 &45.8 &49.4 &59.3\\
&JA-Net&48.3 &44.2 &52.9 &75.7 &74.3 &82.1 &87.6 &61.0 &39.7 &\textbf{63.1} &42.8 &46.9 &59.9\\
&J$\hat{\text{A}}$A-Net w/o $E_{local\_au}$&48.8 &43.1 &50.6 &77.1 &\textbf{76.6} &81.4 &87.1 &\textbf{64.0} &43.8 &61.0 &46.9 &52.0 &61.0\\
&\textbf{J$\hat{\text{A}}$A-Net}&\textbf{53.8}&\textbf{47.8}&\textbf{58.2}&\textbf{78.5}&75.8&82.7&88.2&63.7&43.3&61.8&45.6&49.9&\textbf{62.4}\\\cline{2-15} &J$\hat{\text{A}}$A-Net ($\xi=0$)&45.6 &45.9 &53.7 &76.3 &73.8 &81.5 &\textbf{88.6} &62.8 &48.8 &62.4 &\textbf{48.7} &\textbf{53.0} &61.8\\
&J$\hat{\text{A}}$A-Net ($\hat{p}^{(l)}_i$)&49.7 &41.7 &54.3 &77.0 &75.3 &82.5 &88.3 &60.1 &\textbf{49.4} &58.8 &45.6 &49.4 &61.0\\
\hline
\multirow{12}*{Accuracy}&R-Net&72.2 &79.2 &74.7 &75.3 &64.6 &73.2 &81.3 &60.0 &81.0 &71.8 &80.1 &81.9 &74.6\\
&H-Net&72.3 &81.5 &76.6 &76.0 &65.6 &73.7 &82.0 &59.0 &80.8 &71.5 &79.6 &82.3 &75.1\\
&HD-Net&70.5 &78.9 &75.9 &76.5 &66.3 &77.1 &82.4 &60.7 &80.3 &73.0 &82.6 &85.0 &75.8\\
&HDW-Net&72.4 &79.9 &75.6 &76.6 &67.6 &77.9 &83.3 &60.8 &80.6 &\textbf{74.3} &83.1 &85.0 &76.4\\
&J-Net w/o IF&76.3 &80.4 &78.2 &78.3 &67.5 &77.1 &84.8 &61.3 &81.7 &72.1 &81.2 &85.4 &77.0\\
&J-Net&77.0 &78.9 &78.5 &78.1 &69.8 &76.2 &85.2 &\textbf{65.4} &81.7 &70.5 &82.5 &84.7 &77.4\\
&JA-Net w/o GF&75.9 &78.0 &78.1 &78.3 &72.8 &75.9 &86.5 &62.7 &\textbf{84.3} &70.2 &81.7 &86.0 &77.5\\
&JA-Net&77.4 &77.8 &78.7 &77.7 &71.4 &78.0 &86.4 &64.7 &81.9 &72.8 &83.4 &85.2 &78.0\\
&J$\hat{\text{A}}$A-Net w/o $E_{local\_au}$&\textbf{78.4} &81.0 &77.7 &78.8 &\textbf{74.2} &78.1 &85.9 &63.1 &82.3 &72.9 &81.4 &84.6 &78.2\\
&\textbf{J$\hat{\text{A}}$A-Net}&75.2&80.2&\textbf{82.9}&\textbf{79.8}&72.3&78.2&86.6&65.1&81.0&72.8&82.9&\textbf{86.3}&\textbf{78.6}\\\cline{2-15}
&J$\hat{\text{A}}$A-Net ($\xi=0$)&77.2 &\textbf{81.9} &78.3 &78.7 &72.1 &78.1 &\textbf{87.2} &64.5 &82.8 &72.0 &81.8 &\textbf{86.3} &78.4\\
&J$\hat{\text{A}}$A-Net ($\hat{p}^{(l)}_i$)&73.8 &78.6 &79.7 &78.4 &71.3 &\textbf{78.9} &86.4 &62.9 &83.6 &72.7 &\textbf{83.7} &\textbf{86.3} &78.0\\
\hline
\end{tabular}
\end{table*}

\subsection{Ablation Study}

\subsubsection{Each Component in J$\hat{\text{A}}$A-Net}
\label{sssec:abla_compo}

To investigate the effectiveness of each component in our J$\hat{\text{A}}$A-Net framework, we implement different variants of J$\hat{\text{A}}$A-Net, as summarized in Table~\ref{tab:variant_JAA}. R-Net is a baseline method which is composed by a region learning module with $R(l,l,c)$ and $R(l/2,l/2,2c)$, the global feature learning module, and the two fully-connected layers with dimensions $d$ and $2n_{au}$. Besides, it only employs a multi-label cross entropy loss. J$\hat{\text{A}}$A-Net is named because of \textit{Joint} learning and \textit{Adaptive Attention}. In this way, J-Net only considers \textit{Joint} learning, and JA-Net further uses predefined \textit{Attention} maps to extract local AU features. Table~\ref{tab:ablation_bp4d} presents the results of different variants of J$\hat{\text{A}}$A-Net on BP4D benchmark.

\noindent\textbf{Hierarchical and Multi-Scale Region Layer.}
Comparing the results of H-Net with R-Net, we can observe that our proposed hierarchical and multi-scale region layer improves the performance of AU detection. This is due to that multi-scale partitions help adapt to AUs with different sizes, and hierarchical structure enlarges receptive fields on top of the region layer~\citep{zhao2016deep}. In addition to the stronger feature learning ability, the hierarchical and multi-scale region layer utilizes fewer parameters. As illustrated in Fig.~\ref{fig:multi_region_layer}, except for the common first plain convolutional layer, the parameters of $R(l_1,l_2,c_1)$ is $(3\times3\times4c_1+1)\times4c_1\times8\times8=9216c_1^2+256c_1$, while the parameters of $R_{hm}(l_1,l_2,c_1)$ is $(3\times3\times4c_1+1)\times2c_1\times8\times8+(3\times3\times2c_1+1)\times c_1\times4\times4+(3\times3\times c_1+1)\times c_1\times2\times2=4932c_1^2+148c_1$, where adding $1$ corresponds to the biases of convolutional filters.

\noindent\textbf{Dice Coefficient Loss.}
By integrating the Dice coefficient loss with the cross entropy loss, HD-Net achieves higher average F1-frame and average accuracy than H-Net. This profits from the Dice coefficient loss which optimizes networks from the perspective of the evaluation metric F1-score. The cross entropy loss is effective for classification, but often makes the predictions strongly bias towards non-occurrence for some AUs with low occurrence rates in the training set. The Dice coefficient loss can suppress the prediction bias by focusing on precision and recall which are both related to the true positive.

\noindent\textbf{Weighting the Loss of Each AU.} After using $w_i$ to weight the loss of each AU, HDW-Net improves the average F1-frame and average accuracy to be $57.4$ and $76.4$ over HD-Net, respectively. Benefiting from the weighting to address the data imbalance issue, our method obtains more significant and balanced performance.

\begin{table*}
\centering\caption{F1-frame (F1) and accuracy (Acc) results on BP4D using different adaptive attention learning strategies.}
\label{tab:pre_strategy_bp4d}
\begin{tabular}{|c|c|*{13}{p{0.185in}<{\centering}|}}
\hline
\multicolumn{2}{|c|}{AU}&1&2&4&6&7&10&12&14&15&17&23&24&Avg\\
\hline
\multirow{5}*{F1}&J$\hat{\text{A}}$A-Net w/o $E_{local\_au}$&48.8 &43.1 &50.6 &77.1 &\textbf{76.6} &81.4 &87.1 &\textbf{64.0} &\textbf{43.8} &61.0 &46.9 &52.0 &61.0\\
&J$\hat{\text{A}}$A-Net w/o $E_{local\_au}$+BE&49.6 &47.2 &52.1 &77.5 &74.7 &82.5 &88.0 &61.9 &43.1 &\textbf{63.4} &46.5 &\textbf{52.5} &61.6\\
&J$\hat{\text{A}}$A-Net w/o $E_{local\_au}$+BE+$E_r$&47.1 &46.9 &51.1 &77.1 &76.2 &\textbf{82.8} &88.0 &61.7 &43.2 &63.0 &44.3 &48.1 &60.8\\
&J$\hat{\text{A}}$A-Net w/o $E_{local\_au}$+BE ($\lambda_{e}=3$)&45.7 &46.4 &53.8 &78.3 &74.2 &82.1 &\textbf{88.3} &61.9 &40.4 &62.7 &\textbf{47.3} &51.7 &61.1\\
&\textbf{J$\hat{\text{A}}$A-Net}&\textbf{53.8}&\textbf{47.8}&\textbf{58.2}&\textbf{78.5}&75.8&82.7&88.2&63.7&43.3&61.8&45.6&49.9&\textbf{62.4}\\
\hline
\multirow{5}*{Acc}&J$\hat{\text{A}}$A-Net w/o $E_{local\_au}$&\textbf{78.4} &81.0 &77.7 &78.8 &\textbf{74.2} &78.1 &85.9 &63.1 &82.3 &72.9 &81.4 &84.6 &78.2\\
&J$\hat{\text{A}}$A-Net w/o $E_{local\_au}$+BE&75.7 &\textbf{83.6} &77.5 &79.7 &72.6 &\textbf{78.8} &86.5 &62.5 &\textbf{82.6} &\textbf{73.4} &81.8 &86.3 &78.4\\
&J$\hat{\text{A}}$A-Net w/o $E_{local\_au}$+BE+$E_r$&75.6 &80.4 &76.9 &78.1 &72.4 &78.0 &86.4 &64.2 &81.8 &73.0 &82.7 &86.1 &78.0\\
&J$\hat{\text{A}}$A-Net w/o $E_{local\_au}$+BE ($\lambda_{e}=3$)&77.7 &81.4 &81.5 &78.4 &70.8 &78.0 &\textbf{86.7} &61.4 &81.7 &71.9 &\textbf{84.0} &\textbf{86.4} &78.3\\
&\textbf{J$\hat{\text{A}}$A-Net}&75.2&80.2&\textbf{82.9}&\textbf{79.8}&72.3&78.2&86.6&\textbf{65.1}&81.0&72.8&82.9&86.3&\textbf{78.6}\\
\hline
\end{tabular}
\end{table*}

\noindent\textbf{Integration of Features.}
Compared to HDW-Net, J-Net w/o IF achieves better results by adding the face alignment module, in which the face alignment feature is not fed into AU detection. When integrating the face alignment feature and the global feature for AU detection, J-Net further improves the AU detection performance. These demonstrate that the joint learning with face alignment, \highlight{as well as the face alignment feature with global facial shape information and local landmark information} both contribute to AU detection.

When integrating the two tasks deeper by utilizing the estimated landmarks to generate predefined attention maps, JA-Net w/o GF advances the average F1-frame from level $58$ to level $59$. Specifically, JA-Net w/o GF element-wise multiplies predefined attention maps with the multi-scale feature, and only integrating the face alignment feature and the assembled local features. Its improvement over J-Net shows the effectiveness of the assembled local features. 

Since the global feature can provide \highlight{complementary glo-bal facial structure and texture information}, JA-Net achieves better performance with the average F1-frame of $59.9$ by adding the global feature. However, the predefined attention maps use a fixed size and a fixed attention distribution for each sub-ROI and completely ignore regions beyond the sub-ROIs, which makes JA-Net fail to adapt to AUs with different sizes and exploit correlations among different facial parts.

\noindent\textbf{Adaptive Attention Learning.}
To adapt to various AUs, our J$\hat{\text{A}}$A-Net employs the AU attention refinement step to adaptively learn attention maps, and uses the local AU detection loss $E_{local\_au}$ to supervise the attention refinement. The comparison results of JA-Net, J$\hat{\text{A}}$A-Net w/o $E_{local\_au}$, and J$\hat{\text{A}}$A-Net demonstrate that AU attention refinement and $E_{local\_au}$ are both beneficial for extracting precise local AU features so as to facilitate AU detection.

In J$\hat{\text{A}}$A-Net, each AU sub-ROI is predefined as an approximate Gaussian attention distribution with $\xi=0.56$. To validate the robustness of our proposed adaptive attention learning, we set $\xi=0$ to give predefined attention weight value $1$ for all the points in the sub-ROIs. We can observe that J$\hat{\text{A}}$A-Net ($\xi=0$) achieves comparable performance to J$\hat{\text{A}}$A-Net. Besides, we show the results of $\hat{p}^{(l)}_i$ predicted by the local AU detection loss $E_{local\_au}$ in J$\hat{\text{A}}$A-Net, whose average F1-frame $61.0$ and average accuracy $78.0$ are worse than $62.4$ and $78.6$ of the final predictions $\hat{p}_i$. This again indicates the usefulness of face alignment feature and global feature. 

\subsubsection{Different Strategies for Adaptive Attention Learning}

\noindent\textbf{Back-Propagation Enhancement and Refinement Constraint.}
In the earlier conference version~\citep{shao2018deep}, an alternative strategy for adaptive attention learning is proposed. In particular, to enhance the supervision from $E_{all\_au}$ on the AU attention refinement step, a back-propagation enhancement method is proposed to enlarge the back-propagated gradients for each attention map: 
\begin{equation} \label{eq:Eatt}
\frac{\partial E_{all\_au}}{\partial \hat{\mathbf{V}}_i} \leftarrow \lambda_{e} \frac{\partial E_{all\_au}}{\partial \hat{\mathbf{V}}_i},
\end{equation}
where $\lambda_{e} \ge 1$ is the enhancement coefficient. By enhancing the gradients from $E_{all\_au}$, the attention maps are performed stronger adaptive refinement. The default value of $\lambda_{e}$ is $2$. 

Besides, to avoid the refined attention maps deviating too far from the predefined attention maps, a constraint is introduced for AU attention refinement:
\begin{equation} \label{eq:Er}
E_{r} = - \sum_{i=1}^{n_{au}} \sum_{k=1}^{n_{am}} [v_{ik} \log \hat{v}_{ik} + (1-v_{ik}) \log (1-\hat{v}_{ik})],
\end{equation}
where $n_{am} = l/4 \times l/4$ is the number of points in each refined attention map, and each predefined attention map with size $l_{a_{pre}}\times l_{a_{pre}}\times 1$ is resized to be $l/4\times l/4\times 1$ as the ground truths. Eq.~\eqref{eq:Er} essentially measures the cross entropy between the refined and predefined attention maps.

\noindent\textbf{Discussions.}
The results using the back-propagation enhancement and the refinement constraint $E_r$ are shown in Table~\ref{tab:pre_strategy_bp4d}. J$\hat{\text{A}}$A-Net w/o $E_{local\_au}$+BE+$E_r$ is the re-implementation of the earlier version~\citep{shao2018deep}, \highlight{in which the only differences lie in the removal of cropping attention maps and the replacement of element-wise sum with element-wise average for assembling local feature maps}. We can observe that J$\hat{\text{A}}$A-Net w/o $E_{local\_au}$+BE achieves better performance than J$\hat{\text{A}}$A-Net w/o $E_{local\_au}$, which demonstrates the effectiveness of the back-propagation enhancement. However, after further employing the $E_r$, the results of J$\hat{\text{A}}$A-Net w/o $E_{local\_au}$+BE+$E_r$ become worse. This because $E_r$ enforces a limited solution space for the refinement of attention maps, in which the optimal solutions are often ignored. 

We also notice that J$\hat{\text{A}}$A-Net w/o $E_{local\_au}$+BE ($\lambda_e=3$) fails to obtain better results than J$\hat{\text{A}}$A-Net w/o $E_{local\_au}$+BE when the value of $\lambda_e$ is increased from $2$ to $3$. This indicates that the selection of $\lambda_e$ value is important. In addition, the information of irrelevant AUs from $E_{all\_au}$ may disturb the attention refinement of current AU. Instead, we introduce the local AU detection loss $E_{local\_au}$ \highlight{without the requirement of hyper-parameter $\lambda_e$} to generalize the idea of back-propagation enhancement, in which the influence by irrelevant AUs is also avoided. Besides, we discard the the refinement constraint $E_r$ to learn more precise attention maps. In this way, our J$\hat{\text{A}}$A-Net is more general and achieves better performance than the earlier version~\citep{shao2018deep}.

\begin{figure*}
\centering\includegraphics[width=\linewidth]{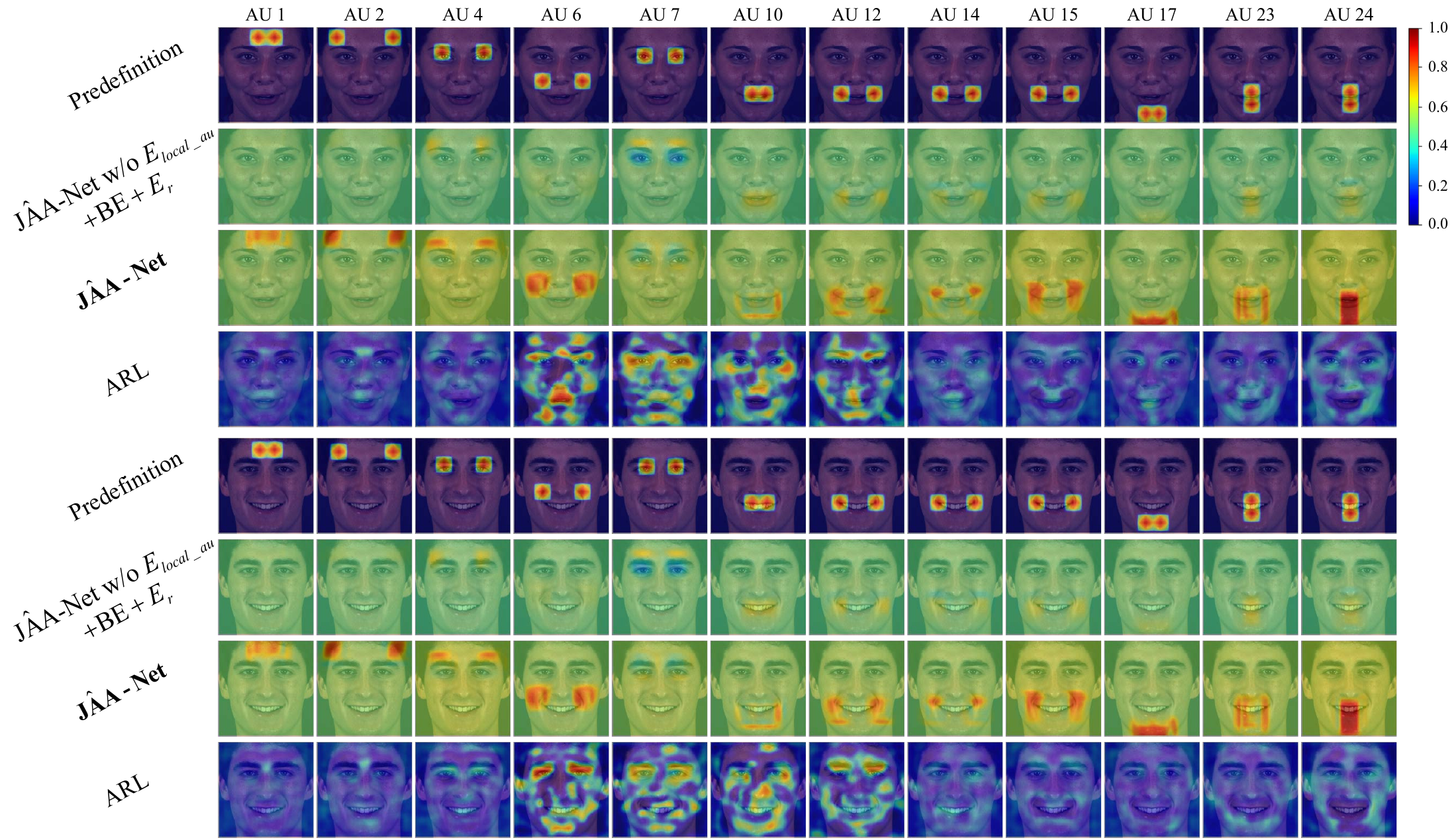}
\caption{Visualization of learned attention maps by different methods for two example BP4D images. The first and fifth rows show the predefined attention map of each AU. Attention weights are visualized with different colors in the color bar, which are overlaid on the images for a better view.}
\label{fig:attention_map}
\end{figure*}

The learned attention maps by J$\hat{\text{A}}$A-Net w/o $E_{local\_au}$+ BE+$E_r$, J$\hat{\text{A}}$A-Net, and ARL for two example BP4D images are visualized in Fig.~\ref{fig:attention_map}. By the adaptive attention learning, the refined attention maps of J$\hat{\text{A}}$A-Net and J$\hat{\text{A}}$A-Net w/o $E_{local\_au}$+BE+$E_r$ both adjust the predefined size and the predefined attention distribution of each AU sub-ROI. We can see that J$\hat{\text{A}}$A-Net adaptively refines the attention maps within a larger range so as to capture more accurate irregular region of each AU than J$\hat{\text{A}}$A-Net w/o $E_{local\_au}$+BE+$E_r$. For example, AUs 12, 14, and 15 with different characteristics have the same predefined attention maps. The refined attention maps of these AUs by J$\hat{\text{A}}$A-Net w/o $E_{local\_au}$+BE+$E_r$ are similar, while their refined attention maps by J$\hat{\text{A}}$A-Net look different. We also notice that there are low attentions in regions beyond the sub-ROIs for the refined attention maps, which contributes to exploiting correlations among different facial parts. With the adaptively refined attention maps, our J$\hat{\text{A}}$A-Net can accurately capture the local feature with respect to each AU.

Fig.~\ref{fig:attention_map} also presents the learned attention maps of a recent method ARL~\citep{shao2019facial}. It learns the attention map of each AU only with the supervision of AU detection. We can see that each learned attention map by ARL captures the rough AU region as well as other correlated regions. However, some irrelevant regions are also captured, which have a bad impact on AU detection. On the other hand, our J$\hat{\text{A}}$A-Net can accurately detect the irregular region of each AU with the help of landmark knowledge, while the regions beyond the irregular sub-ROIs are treated with equal importance. 
Although a few correlated regions far from the sub-ROIs are not highlighted, the use of the face alignment feature and global feature can supplement other required information. Moreover, the comparison results in Sec.~\ref{ssec:comp} also demonstrate better performance of our J$\hat{\text{A}}$A-Net than ARL.

\subsection{J$\hat{\text{A}}$A-Net for Face Alignment}

We have validated the contribution of face alignment to AU detection in Sec.~\ref{sssec:abla_compo}. To also investigate the effectiveness of AU detection for face alignment, we implement a baseline face alignment method called J$\hat{\text{A}}$A-Net w/o AU which only achieves the face alignment task with the removal of AU detection components. Specifically, it consists of the hierarchical and multi-scale region learning module, face alignment module, and two fully-connected layers with dimensions $d$ and $2n_{align}$. 

\begin{table}
\centering\caption{Mean error (lower is better) and failure rate (lower is better) results of different face alignment methods on BP4D.}
\label{tab:align_res}
\begin{tabular}{|*{3}{c|}}
\hline
Method &Mean Error &Failure Rate\\
\hline
\tabincell{c}{ERT\\\citep{kazemi2014one}} &4.73 &3.49 \\
\tabincell{c}{TCDCN\\\citep{zhang2016learning}} &6.57 &1.88\\
\tabincell{c}{MCL\\\citep{shao2019deep}} &7.20 &1.69\\
\tabincell{c}{OpenPose\\\citep{cao2018openpose}} &3.93 &0.27\\
\highlight{\tabincell{c}{LAB\\\citep{wu2018look}}} &4.50 &0.11\\
\highlight{\tabincell{c}{HRNetV2\\\citep{wang2020deep}}} &4.35 &\textbf{0.02}\\\hline
J$\hat{\text{A}}$A-Net w/o AU &6.78 &6.22\\
\textbf{J$\hat{\text{A}}$A-Net} &\textbf{3.80} &0.32\\
\hline
\end{tabular}
\end{table}

Table~\ref{tab:align_res} shows the mean error and failure rate results of J$\hat{\text{A}}$A-Net and J$\hat{\text{A}}$A-Net w/o AU on BP4D benchmark. We also compare with state-of-the-art face alignment methods with trained models released, including ERT~\citep{kazemi2014one}, TCDCN~\citep{zhang2016learning}, MCL~\citep{shao2019deep}, OpenPose~\citep{cao2018openpose,simon2017hand}, \highlight{LAB~\citep{wu2018look}, and HRNetV2~\citep{wang2020deep}}. It can be seen that our J$\hat{\text{A}}$A-Net significantly outperforms ERT, TCDCN and MCL in terms of both mean error and failure rate. \highlight{The recently proposed OpenPose, LAB, and HRNetV2 achieve very low failure rates, while J$\hat{\text{A}}$A-Net has the lowest mean error. Despite being devised for AU detection, our J$\hat{\text{A}}$A-Net's performance is comparable to other methods.}

\begin{figure}
\centering\includegraphics[width=\linewidth]{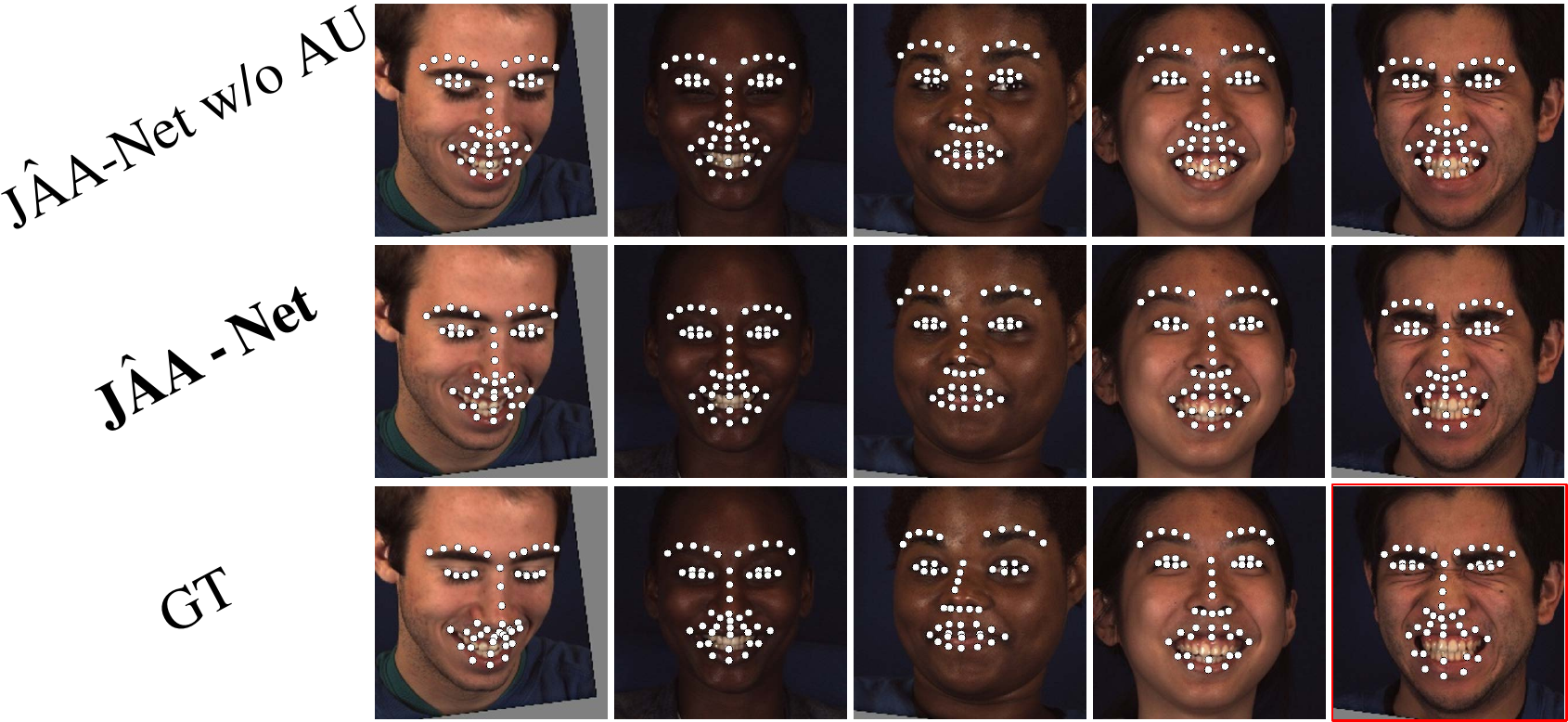}
\caption{Face alignment results for example images from BP4D. ``GT'' denotes the ground-truth locations of facial landmarks.}
\label{fig:land_res}
\end{figure}

Besides, with the help of AU detection, J$\hat{\text{A}}$A-Net outperforms J$\hat{\text{A}}$A-Net w/o AU with large margins of $2.98$ and $5.90$ for mean error and failure rate, respectively. This demonstrates that the AU detection task is also beneficial for face alignment. Face alignment and AU detection contribute to each other in our proposed joint framework. We also compare J$\hat{\text{A}}$A-Net with J$\hat{\text{A}}$A-Net w/o AU on several example images from BP4D in Fig.~\ref{fig:land_res}. We can observe that J$\hat{\text{A}}$A-Net more accurately detects landmarks than J$\hat{\text{A}}$A-Net w/o AU especially for the landmarks in regions of eyes and mouth. In particular, J$\hat{\text{A}}$A-Net shows higher accuracy for eyes of the former three images, as well as mouth of the latter two images. Note that the ground-truth annotations of landmarks along the eyes for the last image (in a red box) are inaccurate, while our J$\hat{\text{A}}$A-Net can detect its eyes well. Moreover, for AU detection, our J$\hat{\text{A}}$A-Net is robust to the minor errors in ground-truth landmark annotations of training data. This is because our J$\hat{\text{A}}$A-Net learns attention maps adaptively, which only requires rough locations of predefined AU centers.

\subsection{J$\hat{\text{A}}$A-Net for Faces with Partial Occlusions}
\label{ssec:occlu}

\begin{figure}
\centering\includegraphics[width=\linewidth]{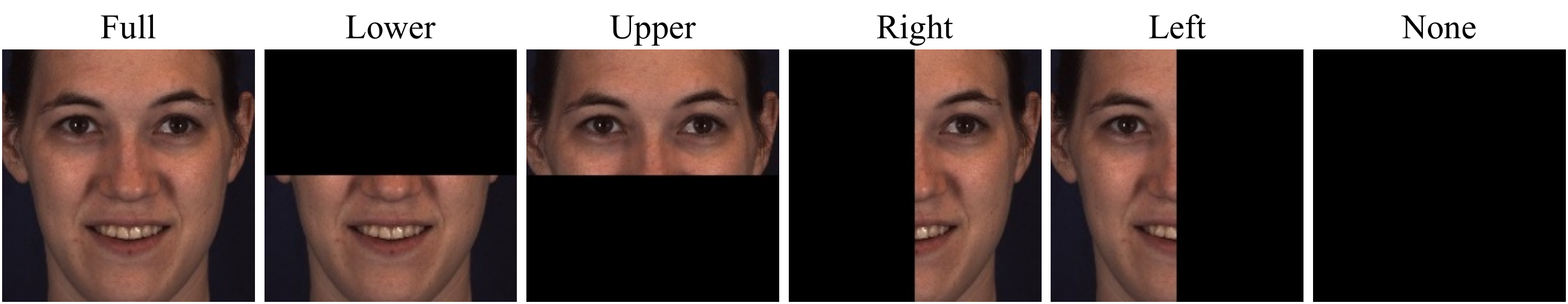}
\caption{An example BP4D face with different occlusions, in which ``Full'' means the whole face is visible.}
\label{fig:occlu_example}
\end{figure}

In this section, we investigate the influences of partial occlusions on AU detection. Following the settings of~\cite{li2018eac,shao2019facial}, we directly employ the trained J$\hat{\text{A}}$A-Net models on BP4D via 3-fold cross-validation to evaluate the corresponding testing set with partial occlusions. Specifically, each testing face is occluded with only lower, upper, right, and left half-faces visible, respectively. Fig.~\ref{fig:occlu_example} illustrates an example face with different partial occlusions. The F1-frame results of EAC-Net~\citep{li2018eac}, ARL~\citep{shao2019facial}, and J$\hat{\text{A}}$A-Net on partially occluded BP4D faces are shown in Table~\ref{tab:f1_occlu}.

\begin{table}
\centering\caption{F1-frame results of our J$\hat{\text{A}}$A-Net and state-of-the-art methods on partially occluded BP4D faces. The best result of each row is shown in bold, and the best average results of all methods are shown in brackets.}
\label{tab:f1_occlu}
\begin{tabular}{|*{7}{c|}}
\hline
\multicolumn{2}{|c|}{AU} &Full &Lower &Upper &Right &Left\\\hline
\multirow{13}*{\rotatebox[origin=c]{90}{EAC-Net~\citep{li2018eac}}}&1 &\textbf{39.0} &31.8 &27.4 &31.4 &25.2\\
&2 &\textbf{35.2} &30.0 &31.6 &34.6 &32.4\\
&4 &\textbf{48.6} &21.8 &29.1 &21.1 &28.7\\
&6 &\textbf{76.1} &70.5 &54.9 &39.7 &52.9\\
&7 &72.9 &72.3 &\textbf{74.4} &66.4 &70.1\\
&10 &\textbf{81.9} &77.0 &64.6 &60.9 &62.6\\
&12 &\textbf{86.2} &75.0 &67.6 &57.9 &59.9\\
&14 &\textbf{58.8} &58.5 &51.6 &48.2 &45.7\\
&15 &\textbf{37.5} &15.0 &14.8 &7.3 &18.4\\
&17 &\textbf{59.1} &58.1 &39.3 &38.7 &37.8\\
&23 &\textbf{35.9} &28.6 &18.9 &27.1 &12.5\\
&24 &\textbf{35.8} &16.3 &13.0 &4.3 &7.6\\
\cline{2-7}&Avg &\textbf{55.9} &[46.3] &40.6 &36.5 &37.8\\
\hline
\multirow{13}*{\rotatebox[origin=c]{90}{ARL~\citep{shao2019facial}}}&1 &\textbf{45.8} &6.3 &20.4 &36.1 &43.3\\
&2 &\textbf{39.8} &8.9 &17.1 &24.2 &38.5\\
&4 &\textbf{55.1} &34.3 &17.1 &43.2 &44.7\\
&6 &\textbf{75.7} &69.3 &57.1 &58.7 &46.8\\
&7 &\textbf{77.2} &70.2 &65.6 &75.4 &59.9\\
&10 &\textbf{82.3} &68.0 &56.3 &57.8 &66.4\\
&12 &\textbf{86.6} &77.2 &68.6 &56.9 &55.3\\
&14 &\textbf{58.8} &53.5 &36.3 &31.4 &35.7\\
&15 &\textbf{47.6} &18.8 &18.0 &0.0 &0.1\\
&17 &\textbf{62.1} &58.8 &51.5 &60.3 &58.3\\
&23 &\textbf{47.4} &30.1 &12.1 &10.2 &26.3\\
&24 &\textbf{55.4} &35.9 &10.4 &34.4 &45.7\\
\cline{2-7}&Avg &\textbf{61.1} &44.3 &35.9 &[40.7] &[43.4]\\
\hline
\multirow{13}*{\rotatebox[origin=c]{90}{\textbf{J$\hat{\text{A}}$A-Net}}}&1 &\textbf{53.8} &6.8 &50.3 &27.4 &37.5\\
&2 &\textbf{47.8} &7.1 &43.8 &31.6 &30.2\\
&4 &\textbf{58.2} &27.6 &36.9 &29.1 &39.7\\
&6 &\textbf{78.5} &67.3 &63.9 &54.9 &65.9\\
&7 &\textbf{75.8} &69.5 &68.6 &74.4 &71.2\\
&10 &\textbf{82.7} &76.4 &74.3 &64.6 &79.8\\
&12 &\textbf{88.2} &79.7 &72.1 &67.6 &76.3\\
&14 &\textbf{63.7} &39.4 &37.4 &51.6 &36.5\\
&15 &\textbf{43.3} &40.1 &17.6 &14.8 &6.5\\
&17 &\textbf{61.8} &58.9 &48.3 &39.3 &23.5\\
&23 &\textbf{45.6} &21.8 &1.7 &18.9 &12.1\\
&24 &\textbf{49.9} &32.4 &20.3 &13.0 &4.6\\
\cline{2-7}&Avg &[\textbf{62.4}] &43.9 &[44.6] &40.2 &40.3\\
\hline
\end{tabular}
\end{table}

\begin{table*}
\centering\caption{Overall F1-frame results on all the $9$ poses of FERA 2017~\citep{valstar2017fera}. CRF~\citep{lafferty2001conditional} is implemented by~\cite{valstar2017fera} as a baseline method.}
\label{tab:f1_bp4d_9_pose}
\begin{tabular}{|*{8}{c|}}
\hline
AU &\rotatebox[origin=c]{45}{\tabincell{c}{CRF\\\citep{lafferty2001conditional}}} &\highlight{\rotatebox[origin=c]{45}{\tabincell{c}{Fusion\\\citep{li2017facial}}}}
&\highlight{\rotatebox[origin=c]{45}{\tabincell{c}{AUMPNet\\\citep{batista2017aumpnet}}}}
&\highlight{\rotatebox[origin=c]{45}{\tabincell{c}{CNN+BLSTM-RNN\\\citep{he2017multi}}}} &\rotatebox[origin=c]{45}{\tabincell{c}{EAC-Net\\\citep{li2018eac}}} &\rotatebox[origin=c]{45}{\tabincell{c}{ARL\\\citep{shao2019facial}}} &\rotatebox[origin=c]{45}{\textbf{J$\hat{\text{A}}$A-Net}}\\\hline
1 &15.4 &28.8 &34.5 &\textbf{36.9} &27.2 &24.0 &30.2\\
4 &17.2 &22.5 &27.8 &26.4 &\textbf{33.2} &28.0 &28.8\\
6 &56.4 &60.0 &67.7 &67.8 &\textbf{69.9} &68.3 &68.2\\
7 &72.7 &74.9 &79.4 &76.3 &\textbf{80.8} &78.1 &78.2\\
10 &69.2 &75.1 &78.5 &80.1 &\textbf{83.4} &75.7 &78.3\\
12 &64.7 &73.0 &76.2 &79.6 &\textbf{80.2} &76.3 &77.8\\
14 &62.2 &60.6 &\textbf{69.2} &66.4 &62.1 &62.7 &67.4\\
15 &14.6 &24.6 &26.7 &26.9 &25.1 &30.0 &\textbf{30.6}\\
17 &22.4 &28.4 &36.4 &36.6 &34.2 &37.9 &\textbf{41.9}\\
23 &20.7 &24.8 &25.0 &24.8 &26.1 &\textbf{39.8} &31.2\\
\hline
Avg &41.6 &47.3 &52.1 &52.2 &52.2 &52.1 &\textbf{53.3}\\
\hline
\end{tabular}
\end{table*}

Compared to EAC-Net and ARL, our J$\hat{\text{A}}$A-Net achieves competitive performance for images with half-faces occluded. In addition, the average F1-frame results of J$\hat{\text{A}}$A-Net for the four half-face occlusions are more balanced than those of EAC-Net and ARL, which demonstrates the robustness of J$\hat{\text{A}}$A-Net on partial occlusions. This is partially due to that our J$\hat{\text{A}}$A-Net jointly trains AU detection and face alignment, and uses predicted landmarks to predefine attention maps, in which the implicit constraint of facial shape contributes to the robustness of landmark detection on occlusions. Besides, compared to ``Full'', the predictions of almost all AUs become worse in half-face images, even though these AUs are not occluded. This indicates that correlations among AUs are beneficial for the detection of a single AU, in which its prediction requires information from correlated AUs in other \highlight{potentially occluded} facial parts.

\begin{figure}
\centering\includegraphics[width=\linewidth]{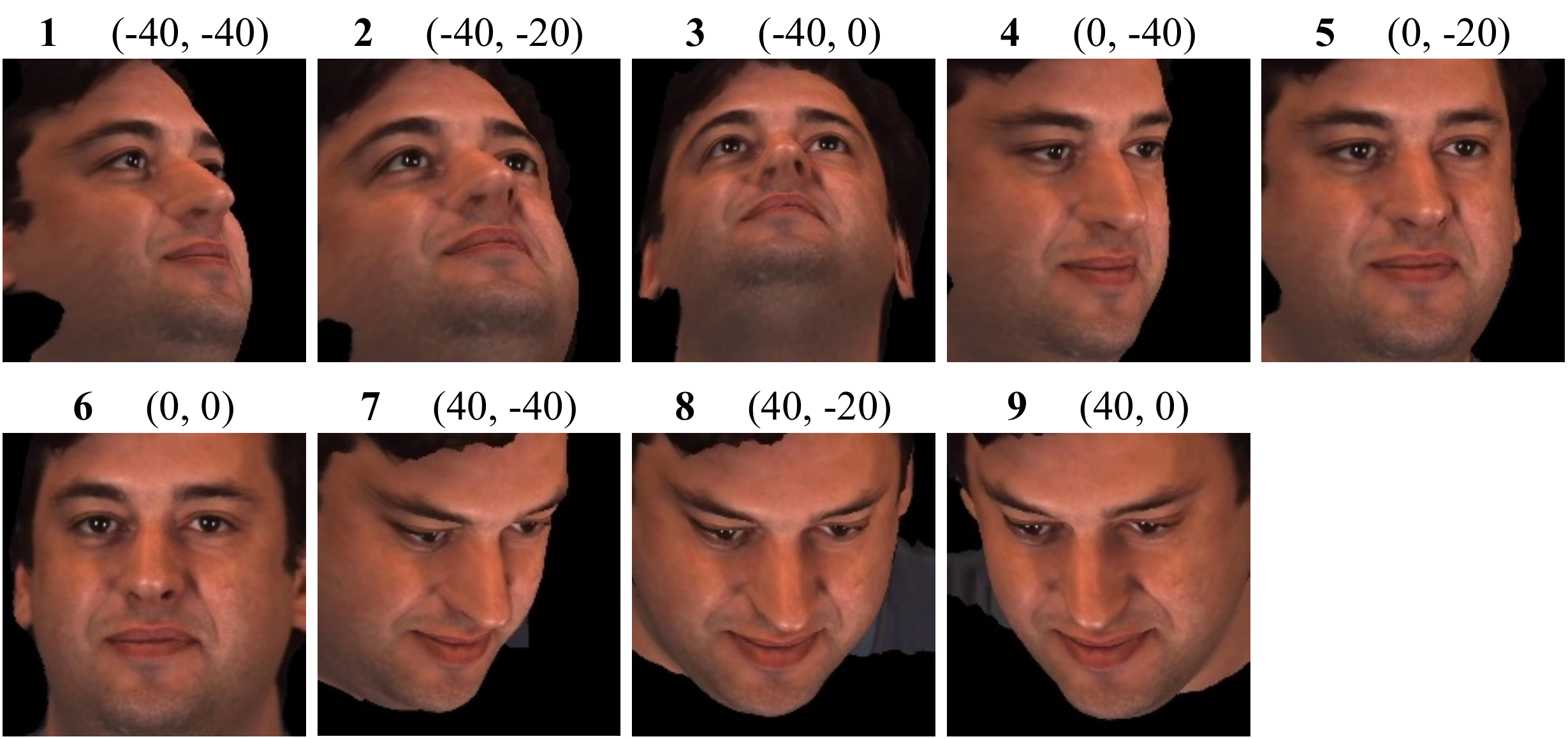}
\caption{An example FERA 2017 face with nine poses, in which the angle degrees (yaw, pitch) of each pose are presented.}
\label{fig:pose_example}
\end{figure}

\subsection{J$\hat{\text{A}}$A-Net for Faces with Non-Frontal Poses}
\label{ssec:non_frontal}

To evaluate our J$\hat{\text{A}}$A-Net on faces with non-frontal head poses, we conduct an experiment on FERA 2017~\citep{valstar2017fera} benchmark. FERA 2017 includes BP4D~\citep{zhang2014bp4d} and BP4D+~\citep{zhang2016multimodal} with $9$ poses, in which each face image is rotated with pitch angles of $-40$, $-20$ and $0$ degrees, and yaw angles of $-40$, $0$ and $40$ degrees from its frontal pose, respectively. The nine poses of an example face is illustrated in Fig.~\ref{fig:pose_example}. Similar to~\cite{li2018eac,shao2019facial}, we utilize BP4D with $41$ subjects for training, a subset of BP4D+ with $20$ subjects for testing, in which 10 AUs (1, 4, 6, 7, 10, 12, 14, 15, 17, and 23) are evaluated. We compare J$\hat{\text{A}}$A-Net against state-of-the-art methods including CRF~\citep{lafferty2001conditional}, \highlight{Fusion~\citep{li2017facial}, AUMPNet~\citep{batista2017aumpnet}, CNN+BLSTM-RNN~\citep{he2017multi},} EAC-Net~\citep{li2018eac}, and ARL~\citep{shao2019facial}.

\begin{table*}
\centering\caption{F1-frame results for each of the $9$ poses of FERA 2017. The best average results of all methods for each pose are shown in bold.}
\label{tab:f1_bp4d_each_pose}
\begin{tabular}{|*{19}{p{0.188in}<{\centering}|}}
\hline
\multirow{2}*{AU} &1&2&3&4&5&6&7&8&9&1&2&3&4&5&6&7&8&9\\
\cline{2-19}&\multicolumn{9}{c|}{CRF~\citep{lafferty2001conditional}} &\multicolumn{9}{c|}{EAC-Net~\citep{li2018eac}}\\\hline
1 &10.3 &15.0 &13.6 &19.3 &19.6 &17.1 &18.0 &14.5 &12.3 &18.8 &37.5 &22.2 &10.1 &40.0 &66.7 &40.1 &33.3 &40.0\\ 
4 &15.0 &15.9 &14.8 &19.1 &19.0 &18.3 &20.2 &20.2 &17.5 &7.0 &40.0 &33.3 &40.0 &30.7 &66.7 &25.0 &18.2 &8.0\\ 
6 &50.5 &55.7 &13.4 &68.9 &74.7 &72.4 &56.0 &53.2 &49.3 &66.7 &66.7 &72.2 &60.8 &68.2 &80.0 &69.2 &75.0 &52.1\\
7 &72.1 &72.9 &41.3 &74.6 &79.7 &78.7 &71.6 &74.7 &75.8 &66.7 &62.9 &73.1 &75.7 &85.7 &85.7 &83.7 &84.7 &73.3\\
10 &55.4 &71.0 &64.2 &77.7 &77.6 &75.0 &63.9 &67.9 &65.9 &85.0 &69.2 &76.9 &83.3 &90.9 &75.0 &62.8 &92.3 &82.8\\
12 &52.2 &67.8 &18.4 &78.6 &80.9 &77.1 &59.6 &63.8 &60.1 &74.3 &68.5 &85.1 &74.2 &86.7 &66.7 &64.3 &84.2 &72.0\\
14 &51.5 &56.3 &9.0 &67.5 &72.4 &74.4 &61.9 &67.0 &67.8 &66.7 &51.1 &51.2 &68.9 &59.3 &50.0 &60.0 &61.2 &74.1\\
15 &13.1 &15.0 &14.2 &14.6 &14.6 &14.6 &14.3 &15.9 &15.2 &57.1 &5.1 &36.4 &25.0 &28.6 &10.2 &7.0 &40.0 &11.8\\
17 &17.3 &25.1 &23.5 &24.2 &24.6 &24.1 &22.0 &21.1 &19.5 &23.5 &36.4 &34.7 &35.3 &46.2 &50.0 &58.8 &21.4 &44.4\\
23 &22.7 &22.9 &19.9 &20.8 &19.6 &16.6 &20.1 &20.1 &20.8 &44.4 &36.3 &7.3 &57.1 &47.1 &6.3 &9.0 &46.2 &11.7\\
Avg &36.0 &41.8 &23.2 &46.5 &48.2 &46.8 &40.8 &41.8 &40.4 &\textbf{50.3} &46.9 &48.5 &52.1 &\textbf{58.3} &54.1 &\textbf{46.4} &\textbf{55.7} &46.2\\
\hline
AU &\multicolumn{9}{c|}{ARL~\citep{shao2019facial}} &\multicolumn{9}{c|}{\textbf{J$\hat{\text{A}}$A-Net}} \\\hline
1 &22.9 &26.3 &24.2 &25.3 &26.3 &26.3 &16.9 &21.0 &24.0 &24.8 &30.7 &28.1 &30.2 &34.5 &34.7 &22.2 &30.9 &31.2\\ 
4 &23.1 &24.6 &29.2 &29.6 &30.0 &30.7 &22.7 &31.3 &30.8 &13.1 &26.3 &22.5 &38.6 &36.5 &34.7 &26.7 &28.2 &28.1\\ 
6 &59.8 &68.0 &71.5 &70.6 &70.0 &71.5 &57.9 &69.8 &70.5 &57.9 &69.0 &70.1 &72.0 &72.3 &72.2 &56.9 &69.6 &68.1\\
7 &70.7 &79.5 &80.8 &79.6 &80.5 &81.0 &72.7 &76.2 &78.2 &75.0 &77.1 &79.2 &78.5 &80.2 &80.9 &74.6 &78.0 &79.7\\
10 &65.4 &78.4 &78.2 &78.8 &79.1 &79.2 &66.4 &75.3 &75.4 &69.2 &79.4 &81.3 &80.5 &81.8 &80.6 &72.0 &78.0 &78.5\\
12 &67.2 &78.1 &78.6 &79.5 &80.3 &81.3 &64.8 &76.1 &76.2 &58.6 &80.0 &80.5 &81.5 &81.8 &83.0 &67.9 &78.9 &79.4\\
14 &58.5 &63.6 &60.9 &62.0 &64.4 &62.8 &58.0 &65.0 &67.2 &64.0 &63.5 &65.3 &69.2 &70.0 &68.8 &67.9 &68.7 &67.2\\
15 &13.9 &31.7 &36.4 &31.9 &34.1 &35.1 &12.3 &30.3 &32.8 &16.1 &27.2 &37.0 &34.8 &32.8 &33.1 &19.6 &33.2 &31.2\\
17 &34.7 &40.4 &40.9 &39.8 &39.9 &40.7 &29.5 &34.6 &36.2 &29.6 &44.0 &43.1 &49.1 &46.5 &43.7 &32.6 &44.0 &41.2\\
23 &28.0 &40.8 &40.3 &44.8 &46.2 &43.7 &31.0 &39.1 &37.4 &22.3 &35.0 &35.6 &35.4 &37.5 &37.1 &17.3 &26.9 &26.7\\
Avg &44.4 &53.1 &54.1 &54.2 &55.1 &55.2 &43.2 &51.9 &52.9 &43.1 &\textbf{53.2} &\textbf{54.3} &\textbf{57.0} &57.4 &\textbf{56.9} &45.8 &53.6 &\textbf{53.1}\\
\hline
\end{tabular}
\end{table*}

Table~\ref{tab:f1_bp4d_9_pose} shows the overall F1-frame results on all the $9$ poses. We can see that our J$\hat{\text{A}}$A-Net outperforms all the other methods. Since facial shape formed by landmarks is related to head poses, the joint learning with face alignment in J$\hat{\text{A}}$A-Net contributes to AU detection with non-frontal poses. \highlight{CNN+BLSTM-RNN is the state-of-the-art method in FERA 2017 challenge, which uses Bidirectional Long Short-Term Memory Recurrent Neural Networks (BLSTM-RNN) to model temporal information among frames. In contrast, J$\hat{\text{A}}$A-Net achieves better performance only based on single input frame.} 

We also present the F1-frame results of CRF, EAC-Net, ARL and J$\hat{\text{A}}$A-Net for each pose in Table~\ref{tab:f1_bp4d_each_pose}. It can be observed that our J$\hat{\text{A}}$A-Net achieves the best performance in terms of average F1-frame for most poses (2, 3, 4, 6, and 9). This demonstrates that our method is robust to faces with various non-frontal poses.

\section{Conclusion}

In this paper, we have developed a novel deep learning framework for joint facial AU detection and face alignment. Joint learning of the two tasks contributes to each other by sharing features and initializing the attention maps with the face alignment results. In addition, we have proposed the adaptive attention learning module to localize irregular regions of AUs adaptively so as to extract more precise local features. Our framework is end-to-end without any post-processing operation.

We have compared our approach with state-of-the-art methods on the challenging BP4D, DISFA, GFT, and BP4D+ benchmarks. It is demonstrated that our approach significantly outperforms the state-of-the-art AU detection methods. In addition, we have conducted an ablation study which indicates that each component in our framework is beneficial for AU detection, and the introduced local AU detection loss is an effective strategy for adaptive attention learning. Besides, the visual results demonstrate that our approach can adaptively capture the irregular region of each AU.

We have further compared our approach against a baseline method with the removal of AU detection components, as well as state-of-the-art face alignment methods. The results indicate that AU detection also contributes to face alignment, and our approach achieves competitive face alignment performance. Moreover, we have conducted experiments to validate the effectiveness and robustness of our framework on faces with partial occlusions and non-frontal poses, respectively. Our proposed framework is also promising to be applied for other face analysis tasks and other multi-task problems.

\begin{acknowledgements}
This work is supported by the Start-up Grant, School of Computer Science and Technology, China University of Mining and Technology. It is also partially supported by the National Key R\&D Program of China (No. 2019YFC1521104), the National Natural Science Foundation of China (No. 61972157 and No. 61503277), the Zhejiang Lab (No. 2020NB0AB01), the Data Science \& Artificial Intelligence Research Centre@NTU (DSAIR), and the Monash FIT Start-up Grant.
\end{acknowledgements}

%
%

\bibliographystyle{spbasic}      
\bibliography{references}   

\begin{thebibliography}{57}
\providecommand{\natexlab}[1]{#1}
\providecommand{\url}[1]{{#1}}
\providecommand{\urlprefix}{URL }
\expandafter\ifx\csname urlstyle\endcsname\relax
  \providecommand{\doi}[1]{DOI~\discretionary{}{}{}#1}\else
  \providecommand{\doi}{DOI~\discretionary{}{}{}\begingroup
  \urlstyle{rm}\Url}\fi
\providecommand{\eprint}[2][]{\url{#2}}

\bibitem[{Batista et~al.(2017)Batista, Albiero, Bellon, and
  Silva}]{batista2017aumpnet}
Batista JC, Albiero V, Bellon OR, Silva L (2017) Aumpnet: Simultaneous action
  units detection and intensity estimation on multipose facial images using a
  single convolutional neural network. In: IEEE International Conference on
  Automatic Face \& Gesture Recognition, IEEE, pp 866--871

\bibitem[{Benitez-Quiroz et~al.(2016)Benitez-Quiroz, Srinivasan, Martinez
  et~al.}]{benitez2016emotionet}
Benitez-Quiroz CF, Srinivasan R, Martinez AM, et~al. (2016) Emotionet: An
  accurate, real-time algorithm for the automatic annotation of a million
  facial expressions in the wild. In: IEEE Conference on Computer Vision and
  Pattern Recognition, IEEE, pp 5562--5570

\bibitem[{Cao et~al.(2019)Cao, Hidalgo, Simon, Wei, and
  Sheikh}]{cao2018openpose}
Cao Z, Hidalgo G, Simon T, Wei SE, Sheikh Y (2019) Openpose: Realtime
  multi-person 2d pose estimation using part affinity fields. IEEE Transactions
  on Pattern Analysis and Machine Intelligence

\bibitem[{Chu et~al.(2017)Chu, De~la Torre, and Cohn}]{chu2017learning}
Chu WS, De~la Torre F, Cohn JF (2017) Learning spatial and temporal cues for
  multi-label facial action unit detection. In: IEEE International Conference
  on Automatic Face \& Gesture Recognition, IEEE, pp 25--32

\bibitem[{Cootes et~al.(2001)Cootes, Edwards, and Taylor}]{cootes2001active}
Cootes TF, Edwards GJ, Taylor CJ (2001) Active appearance models. IEEE
  Transactions on Pattern Analysis and Machine Intelligence 23(6):681--685

\bibitem[{Corneanu et~al.(2016)Corneanu, Sim{\'o}n, Cohn, and
  Guerrero}]{corneanu2016survey}
Corneanu CA, Sim{\'o}n MO, Cohn JF, Guerrero SE (2016) Survey on rgb, 3d,
  thermal, and multimodal approaches for facial expression recognition:
  History, trends, and affect-related applications. IEEE Transactions on
  Pattern Analysis and Machine Intelligence 38(8):1548--1568

\bibitem[{Corneanu et~al.(2018)Corneanu, Madadi, and
  Escalera}]{corneanu2018deep}
Corneanu CA, Madadi M, Escalera S (2018) Deep structure inference network for
  facial action unit recognition. In: European Conference on Computer Vision,
  Springer, pp 309--324

\bibitem[{Ekman and Friesen(1978)}]{ekman1978facial}
Ekman P, Friesen WV (1978) Facial action coding system: A technique for the
  measurement of facial movement. Consulting Psychologists Press

\bibitem[{Ekman et~al.(2002)Ekman, Friesen, and Hager}]{ekman2002facial}
Ekman P, Friesen WV, Hager JC (2002) Facial action coding system. Research
  Nexus

\bibitem[{Ertugrul et~al.(2020)Ertugrul, Cohn, Jeni, Zhang, Yin, and
  Ji}]{ertugrul2020crossing}
Ertugrul IO, Cohn JF, Jeni LA, Zhang Z, Yin L, Ji Q (2020) Crossing domains for
  au coding: Perspectives, approaches, and measures. IEEE Transactions on
  Biometrics, Behavior, and Identity Science 2(2):158--171

\bibitem[{Fan et~al.(2008)Fan, Chang, Hsieh, Wang, and Lin}]{fan2008liblinear}
Fan RE, Chang KW, Hsieh CJ, Wang XR, Lin CJ (2008) Liblinear: A library for
  large linear classification. Journal of Machine Learning Research
  9(Aug):1871--1874

\bibitem[{Girard et~al.(2017)Girard, Chu, Jeni, and Cohn}]{girard2017sayette}
Girard JM, Chu WS, Jeni LA, Cohn JF (2017) Sayette group formation task (gft)
  spontaneous facial expression database. In: IEEE International Conference on
  Automatic Face \& Gesture Recognition, IEEE, pp 581--588

\bibitem[{Gudi et~al.(2015)Gudi, Tasli, Den~Uyl, and Maroulis}]{gudi2015deep}
Gudi A, Tasli HE, Den~Uyl TM, Maroulis A (2015) Deep learning based facs action
  unit occurrence and intensity estimation. In: IEEE International Conference
  and Workshops on Automatic Face and Gesture Recognition, IEEE, vol~6, pp 1--5

\bibitem[{He et~al.(2017)He, Li, Yang, Cao, Sun, and Yu}]{he2017multi}
He J, Li D, Yang B, Cao S, Sun B, Yu L (2017) Multi view facial action unit
  detection based on cnn and blstm-rnn. In: IEEE International Conference on
  Automatic Face \& Gesture Recognition, IEEE, pp 848--853

\bibitem[{He et~al.(2016)He, Zhang, Ren, and Sun}]{he2016deep}
He K, Zhang X, Ren S, Sun J (2016) Deep residual learning for image
  recognition. In: IEEE Conference on Computer Vision and Pattern Recognition,
  IEEE, pp 770--778

\bibitem[{Ioffe and Szegedy(2015)}]{ioffe2015batch}
Ioffe S, Szegedy C (2015) Batch normalization: Accelerating deep network
  training by reducing internal covariate shift. In: International Conference
  on Machine Learning, pp 448--456

\bibitem[{Jeni et~al.(2017)Jeni, Cohn, and Kanade}]{jeni2017dense}
Jeni LA, Cohn JF, Kanade T (2017) Dense 3d face alignment from 2d video for
  real-time use. Image and Vision Computing 58:13--24

\bibitem[{Kazemi and Sullivan(2014)}]{kazemi2014one}
Kazemi V, Sullivan J (2014) One millisecond face alignment with an ensemble of
  regression trees. In: IEEE Conference on Computer Vision and Pattern
  Recognition, IEEE, pp 1867--1874

\bibitem[{Krizhevsky et~al.(2012)Krizhevsky, Sutskever, and
  Hinton}]{krizhevsky2012imagenet}
Krizhevsky A, Sutskever I, Hinton GE (2012) Imagenet classification with deep
  convolutional neural networks. In: Advances in Neural Information Processing
  Systems, Curran Associates, Inc., pp 1097--1105

\bibitem[{Lafferty et~al.(2001)Lafferty, McCallum, and
  Pereira}]{lafferty2001conditional}
Lafferty J, McCallum A, Pereira FC (2001) Conditional random fields:
  Probabilistic models for segmenting and labeling sequence data. In:
  International Conference on Machine Learning, pp 282--289

\bibitem[{Li et~al.(2017{\natexlab{a}})Li, Abtahi, and Zhu}]{li2017action}
Li W, Abtahi F, Zhu Z (2017{\natexlab{a}}) Action unit detection with region
  adaptation, multi-labeling learning and optimal temporal fusing. In: IEEE
  Conference on Computer Vision and Pattern Recognition, IEEE, pp 6766--6775

\bibitem[{Li et~al.(2018)Li, Abtahi, Zhu, and Yin}]{li2018eac}
Li W, Abtahi F, Zhu Z, Yin L (2018) Eac-net: Deep nets with enhancing and
  cropping for facial action unit detection. IEEE Transactions on Pattern
  Analysis and Machine Intelligence 40(11):2583--2596

\bibitem[{Li et~al.(2017{\natexlab{b}})Li, Chen, and Jin}]{li2017facial}
Li X, Chen S, Jin Q (2017{\natexlab{b}}) Facial action units detection with
  multi-features and -aus fusion. In: IEEE International Conference on
  Automatic Face \& Gesture Recognition, IEEE, pp 860--865

\bibitem[{Li et~al.(2013)Li, Wang, Zhao, and Ji}]{li2013simultaneous}
Li Y, Wang S, Zhao Y, Ji Q (2013) Simultaneous facial feature tracking and
  facial expression recognition. IEEE Transactions on Image Processing
  22(7):2559--2573

\bibitem[{Li et~al.(2019)Li, Zeng, Shan, and Chen}]{li2019self-supervised}
Li Y, Zeng J, Shan S, Chen X (2019) Self-supervised representation learning
  from videos for facial action unit detection. In: IEEE Conference on Computer
  Vision and Pattern Recognition, IEEE, pp 10924--10933

\bibitem[{Martinez et~al.(2019)Martinez, Valstar, Jiang, and
  Pantic}]{TAC2017Pantic}
Martinez B, Valstar MF, Jiang B, Pantic M (2019) Automatic analysis of facial
  actions: A survey. IEEE Transactions on Affective Computing 10(3):325--347

\bibitem[{Mavadati et~al.(2013)Mavadati, Mahoor, Bartlett, Trinh, and
  Cohn}]{mavadati2013disfa}
Mavadati SM, Mahoor MH, Bartlett K, Trinh P, Cohn JF (2013) Disfa: A
  spontaneous facial action intensity database. IEEE Transactions on Affective
  Computing 4(2):151--160

\bibitem[{Milletari et~al.(2016)Milletari, Navab, and Ahmadi}]{milletari2016v}
Milletari F, Navab N, Ahmadi SA (2016) V-net: Fully convolutional neural
  networks for volumetric medical image segmentation. In: International
  Conference on 3D Vision, IEEE, pp 565--571

\bibitem[{Nair and Hinton(2010)}]{nair2010rectified}
Nair V, Hinton GE (2010) Rectified linear units improve restricted boltzmann
  machines. In: International Conference on Machine Learning, pp 807--814

\bibitem[{Niu et~al.(2019)Niu, Han, Yang, Huang, and Shan}]{niu2019local}
Niu X, Han H, Yang S, Huang Y, Shan S (2019) Local relationship learning with
  person-specific shape regularization for facial action unit detection. In:
  IEEE Conference on Computer Vision and Pattern Recognition, pp 11917--11926

\bibitem[{Paszke et~al.(2019)Paszke, Gross, Massa, Lerer, Bradbury, Chanan,
  Killeen, Lin, Gimelshein, Antiga, Desmaison, Kopf, Yang, DeVito, Raison,
  Tejani, Chilamkurthy, Steiner, Fang, Bai, and Chintala}]{paszke2019pytorch}
Paszke A, Gross S, Massa F, Lerer A, Bradbury J, Chanan G, Killeen T, Lin Z,
  Gimelshein N, Antiga L, Desmaison A, Kopf A, Yang E, DeVito Z, Raison M,
  Tejani A, Chilamkurthy S, Steiner B, Fang L, Bai J, Chintala S (2019)
  Pytorch: An imperative style, high-performance deep learning library. In:
  Advances in Neural Information Processing Systems, Curran Associates, Inc.,
  pp 8024--8035

\bibitem[{Ranjan et~al.(2019)Ranjan, Patel, and
  Chellappa}]{ranjan2017hyperface}
Ranjan R, Patel VM, Chellappa R (2019) Hyperface: A deep multi-task learning
  framework for face detection, landmark localization, pose estimation, and
  gender recognition. IEEE Transactions on Pattern Analysis and Machine
  Intelligence 41(1):121--135

\bibitem[{Sanchez et~al.(2018)Sanchez, Tzimiropoulos, and
  Valstar}]{sanchez2018joint}
Sanchez E, Tzimiropoulos G, Valstar M (2018) Joint action unit localisation and
  intensity estimation through heatmap regression. In: British Machine Vision
  Conference, BMVA Press, p 233

\bibitem[{Sankaran et~al.(2019)Sankaran, Mohan, Setlur, Govindaraju, and
  Fedorishin}]{sankaran2019representation}
Sankaran N, Mohan DD, Setlur S, Govindaraju V, Fedorishin D (2019)
  Representation learning through cross-modality supervision. In: IEEE
  International Conference on Automatic Face \& Gesture Recognition, IEEE, pp
  1--8

\bibitem[{Shao et~al.(2016)Shao, Ding, Zhao, Zhang, and Ma}]{shao2016learning}
Shao Z, Ding S, Zhao Y, Zhang Q, Ma L (2016) Learning deep representation from
  coarse to fine for face alignment. In: IEEE International Conference on
  Multimedia and Expo, IEEE, pp 1--6

\bibitem[{Shao et~al.(2018)Shao, Liu, Cai, and Ma}]{shao2018deep}
Shao Z, Liu Z, Cai J, Ma L (2018) Deep adaptive attention for joint facial
  action unit detection and face alignment. In: European Conference on Computer
  Vision, Springer, pp 725--740

\bibitem[{Shao et~al.(2019)Shao, Liu, Cai, Wu, and Ma}]{shao2019facial}
Shao Z, Liu Z, Cai J, Wu Y, Ma L (2019) Facial action unit detection using
  attention and relation learning. IEEE Transactions on Affective Computing

\bibitem[{Shao et~al.(2020)Shao, Zhu, Tan, Hao, and Ma}]{shao2019deep}
Shao Z, Zhu H, Tan X, Hao Y, Ma L (2020) Deep multi-center learning for face
  alignment. Neurocomputing 396:477--486

\bibitem[{Simon et~al.(2017)Simon, Joo, Matthews, and Sheikh}]{simon2017hand}
Simon T, Joo H, Matthews I, Sheikh Y (2017) Hand keypoint detection in single
  images using multiview bootstrapping. In: IEEE Conference on Computer Vision
  and Pattern Recognition, IEEE, pp 4645--4653

\bibitem[{Sutskever et~al.(2013)Sutskever, Martens, Dahl, and
  Hinton}]{sutskever2013importance}
Sutskever I, Martens J, Dahl G, Hinton G (2013) On the importance of
  initialization and momentum in deep learning. In: International Conference on
  Machine Learning, pp 1139--1147

\bibitem[{De~la Torre et~al.(2015)De~la Torre, Chu, Xiong, Vicente, Ding, and
  Cohn}]{de2015intraface}
De~la Torre F, Chu WS, Xiong X, Vicente F, Ding X, Cohn JF (2015) Intraface.
  In: IEEE International Conference and Workshops on Automatic Face and Gesture
  Recognition, IEEE, pp 1--8

\bibitem[{Valstar and Pantic(2006)}]{valstar2006fully}
Valstar M, Pantic M (2006) Fully automatic facial action unit detection and
  temporal analysis. In: IEEE Conference on Computer Vision and Pattern
  Recognition Workshop, IEEE, pp 149--149

\bibitem[{Valstar et~al.(2017)Valstar, S{\'a}nchez-Lozano, Cohn, Jeni, Girard,
  Zhang, Yin, and Pantic}]{valstar2017fera}
Valstar MF, S{\'a}nchez-Lozano E, Cohn JF, Jeni LA, Girard JM, Zhang Z, Yin L,
  Pantic M (2017) Fera 2017-addressing head pose in the third facial expression
  recognition and analysis challenge. In: IEEE International Conference on
  Automatic Face \& Gesture Recognition, IEEE, pp 839--847

\bibitem[{Wang et~al.(2020)Wang, Sun, Cheng, Jiang, Deng, Zhao, Liu, Mu, Tan,
  Wang et~al.}]{wang2020deep}
Wang J, Sun K, Cheng T, Jiang B, Deng C, Zhao Y, Liu D, Mu Y, Tan M, Wang X,
  et~al. (2020) Deep high-resolution representation learning for visual
  recognition. IEEE Transactions on Pattern Analysis and Machine Intelligence

\bibitem[{Wu et~al.(2018)Wu, Qian, Yang, Wang, Cai, and Zhou}]{wu2018look}
Wu W, Qian C, Yang S, Wang Q, Cai Y, Zhou Q (2018) Look at boundary: A
  boundary-aware face alignment algorithm. In: IEEE conference on computer
  vision and pattern recognition, IEEE, pp 2129--2138

\bibitem[{Wu and Ji(2016)}]{wu2016constrained}
Wu Y, Ji Q (2016) Constrained joint cascade regression framework for
  simultaneous facial action unit recognition and facial landmark detection.
  In: IEEE Conference on Computer Vision and Pattern Recognition, IEEE, pp
  3400--3408

\bibitem[{Wu et~al.(2017)Wu, Gou, and Ji}]{wu2017simultaneous}
Wu Y, Gou C, Ji Q (2017) Simultaneous facial landmark detection, pose and
  deformation estimation under facial occlusion. In: IEEE Conference on
  Computer Vision and Pattern Recognition, IEEE, pp 3471--3480

\bibitem[{Xiong and De~la Torre(2013)}]{xiong2013supervised}
Xiong X, De~la Torre F (2013) Supervised descent method and its applications to
  face alignment. In: IEEE Conference on Computer Vision and Pattern
  Recognition, IEEE, pp 532--539

\bibitem[{Zeng et~al.(2015)Zeng, Chu, De~la Torre, Cohn, and
  Xiong}]{zeng2015confidence}
Zeng J, Chu WS, De~la Torre F, Cohn JF, Xiong Z (2015) Confidence preserving
  machine for facial action unit detection. In: IEEE International Conference
  on Computer Vision, IEEE, pp 3622--3630

\bibitem[{Zhang et~al.(2014{\natexlab{a}})Zhang, Yin, Cohn, Canavan, Reale,
  Horowitz, Liu, and Girard}]{zhang2014bp4d}
Zhang X, Yin L, Cohn JF, Canavan S, Reale M, Horowitz A, Liu P, Girard JM
  (2014{\natexlab{a}}) Bp4d-spontaneous: A high-resolution spontaneous 3d
  dynamic facial expression database. Image and Vision Computing
  32(10):692--706

\bibitem[{Zhang et~al.(2014{\natexlab{b}})Zhang, Luo, Loy, and
  Tang}]{zhang2014facial}
Zhang Z, Luo P, Loy CC, Tang X (2014{\natexlab{b}}) Facial landmark detection
  by deep multi-task learning. In: European Conference on Computer Vision,
  Springer, pp 94--108

\bibitem[{Zhang et~al.(2016{\natexlab{a}})Zhang, Girard, Wu, Zhang, Liu,
  Ciftci, Canavan, Reale, Horowitz, Yang, Cohn, Ji, and
  Yin}]{zhang2016multimodal}
Zhang Z, Girard JM, Wu Y, Zhang X, Liu P, Ciftci U, Canavan S, Reale M,
  Horowitz A, Yang H, Cohn JF, Ji Q, Yin L (2016{\natexlab{a}}) Multimodal
  spontaneous emotion corpus for human behavior analysis. In: IEEE Conference
  on Computer Vision and Pattern Recognition, IEEE, pp 3438--3446

\bibitem[{Zhang et~al.(2016{\natexlab{b}})Zhang, Luo, Loy, and
  Tang}]{zhang2016learning}
Zhang Z, Luo P, Loy CC, Tang X (2016{\natexlab{b}}) Learning deep
  representation for face alignment with auxiliary attributes. IEEE
  Transactions on Pattern Analysis and Machine Intelligence 38(5):918--930

\bibitem[{Zhao et~al.(2015)Zhao, Chu, De~la Torre, Cohn, and
  Zhang}]{zhao2015joint}
Zhao K, Chu WS, De~la Torre F, Cohn JF, Zhang H (2015) Joint patch and
  multi-label learning for facial action unit detection. In: IEEE Conference on
  Computer Vision and Pattern Recognition, IEEE, pp 2207--2216

\bibitem[{Zhao et~al.(2016{\natexlab{a}})Zhao, Chu, De~la Torre, Cohn, and
  Zhang}]{zhao2016joint}
Zhao K, Chu WS, De~la Torre F, Cohn JF, Zhang H (2016{\natexlab{a}}) Joint
  patch and multi-label learning for facial action unit and holistic expression
  recognition. IEEE Transactions on Image Processing 25(8):3931--3946

\bibitem[{Zhao et~al.(2016{\natexlab{b}})Zhao, Chu, and Zhang}]{zhao2016deep}
Zhao K, Chu WS, Zhang H (2016{\natexlab{b}}) Deep region and multi-label
  learning for facial action unit detection. In: IEEE Conference on Computer
  Vision and Pattern Recognition, IEEE, pp 3391--3399

\bibitem[{Zhong et~al.(2015)Zhong, Liu, Yang, Huang, and
  Metaxas}]{zhong2015learning}
Zhong L, Liu Q, Yang P, Huang J, Metaxas DN (2015) Learning multiscale active
  facial patches for expression analysis. IEEE Transactions on Cybernetics
  45(8):1499--1510

\end{thebibliography}


\end{document}